\documentclass[11pt,a4paper]{article}

\usepackage{authblk}
\usepackage[hyperref]{acl2020}
\usepackage{times}    % uses Times Roman font (DO NOT CHANGE!)
\usepackage{latexsym} % uses Times Roman font (DO NOT CHANGE!)
\usepackage{amsmath}
\usepackage{amssymb}
\usepackage{booktabs}
\usepackage[utf8]{inputenc}
\usepackage{url}
\usepackage{multirow}
\usepackage{array}
\usepackage{graphicx} 
\usepackage{colortbl} 
\usepackage[inline]{enumitem}
\usepackage{bm} 
\usepackage{subcaption} % do not add any optional arguments!
\usepackage{cleveref}
\usepackage{bold-extra} 
\usepackage{empheq}   
\usepackage{amssymb} 
\usepackage{subfiles}
\usepackage{float}

\usepackage[stretch=25,shrink=25]{microtype}

\aclfinalcopy % Uncomment this line for the final submission <<<<<<<<<<<<<<<<<<<<<<<<<<<
\def\aclpaperid{519} %  Enter the acl Paper ID here <<<<<<<<<<<<<<<<<<<<<<<<<<<

\newcommand{\onlyinsubfile}[1]{#1}
\newcommand{\notinsubfile}[1]{}

\newcommand{\norm}[1]{{\lvert #1 \rvert}}

\newcommand{\dG}[1]{\textcolor[RGB]{0,102,0}{#1}}
\newcommand{\dR}[1]{\textcolor[RGB]{153,0,0}{#1}}

\newcommand{\given}[1]{{\mkern1.0mu\vert\mkern1.0mu#1}}
\newcommand{\doublevert}[1]{{\mkern1.0mu\mathbin{\|}\mkern1.0mu#1}}
\newcommand{\posdiff}[1]{{\small(\dG{+#1})}}
\newcommand{\negdiff}[1]{{\small(\dR{-#1})}}
\newcommand{\myeq}{{\mkern3.0mu{=}\mkern3.0mu}}
\newcommand{\myneq}{{\mkern3.0mu{\neq}\mkern3.0mu}}
\newcommand{\meanstddev}[2]{{#1$\pm$#2}}

\newcolumntype{L}[1]{>{\raggedright\arraybackslash} m{#1\textwidth}}
\newcolumntype{R}[1]{>{\raggedleft\arraybackslash} m{#1\textwidth}}
\newcolumntype{C}[1]{>{\centering\arraybackslash} m{#1\textwidth}}
\newcolumntype{X}[1]{>{\raggedright\arraybackslash} m{#1\columnwidth}}
\newcolumntype{Y}[1]{>{\raggedleft\arraybackslash} m{#1\columnwidth}}
\newcolumntype{Z}[1]{>{\centering\arraybackslash} m{#1\columnwidth}}

\title{
NAT: Noise-Aware Training for Robust Neural Sequence Labeling
}

\author[1,2]{{\bf Marcin Namysl}}
\author[1,2]{{\bf Sven Behnke}}
\author[1]{{\bf Joachim Köhler}}

\affil[1]{
Fraunhofer IAIS\\
Sankt Augustin, Germany
}

\affil[2]{
Autonomous Intelligent Systems\\
Computer Science Institute VI\\
University of Bonn, Germany
}

\affil[ ]{
{\tt \{Marcin.Namysl, Sven.Behnke, Joachim Koehler\}@iais.fraunhofer.de}
}

\date{}

\begin{document}

\renewcommand{\onlyinsubfile}[1]{}
\renewcommand{\notinsubfile}[1]{#1}

\maketitle

\begin{abstract}
Sequence labeling systems should perform reliably not only under ideal conditions but also with corrupted inputs---as these systems often process user-generated text or follow an error-prone upstream component.
To this end, we formulate the noisy sequence labeling problem, where the input may undergo an unknown noising process and 
propose two Noise-Aware Training (NAT) objectives that improve robustness of sequence labeling performed on perturbed input:
Our data augmentation method trains a neural model using a mixture of clean and noisy samples, whereas our stability training algorithm encourages the model to create a noise-invariant latent representation.
We employ a vanilla noise model at training time.
For evaluation, we use both the original data and its variants perturbed with real OCR errors and misspellings.
Extensive experiments on English and German named entity recognition benchmarks confirmed that NAT consistently improved robustness of popular sequence labeling models, preserving accuracy on the original input.
We make our code and data publicly available for the research community.

\end{abstract}

\section{Introduction}
\label{sec:intro}

Sequence labeling systems are generally trained on clean text, although in real-world scenarios, they often follow an error-prone upstream component, such as Optical Character Recognition (OCR;~\citealp{neudecker-2016-open}) or Automatic Speech Recognition (ASR;~\citealp{DBLP:conf/interspeech/ParadaDJ11}). 
Sequence labeling is also often performed on user-generated text, which may contain spelling mistakes or typos~\citep{derczynski-etal-2013-twitter}.
Errors introduced in an upstream task are propagated downstream, diminishing the performance of the end-to-end system~\citep{Alex:2014:ERQ:2595188.2595214}.
While humans can easily cope with typos, misspellings, and the complete omission of letters when reading~\citep{4126417}, most Natural Language Processing (NLP) systems fail when processing corrupted or noisy text~\citep{DBLP:conf/iclr/BelinkovB18}. Although this problem is not new to NLP, only a few works addressed it explicitly~\citep{piktus-etal-2019-misspelling,karpukhin-etal-2019-training}. Other methods must rely on the noise that occurs naturally in the training data.

\begin{figure}[t]
%\begin{figure}[!htbp]
\begin{center}
\includegraphics[width=1.0\columnwidth]{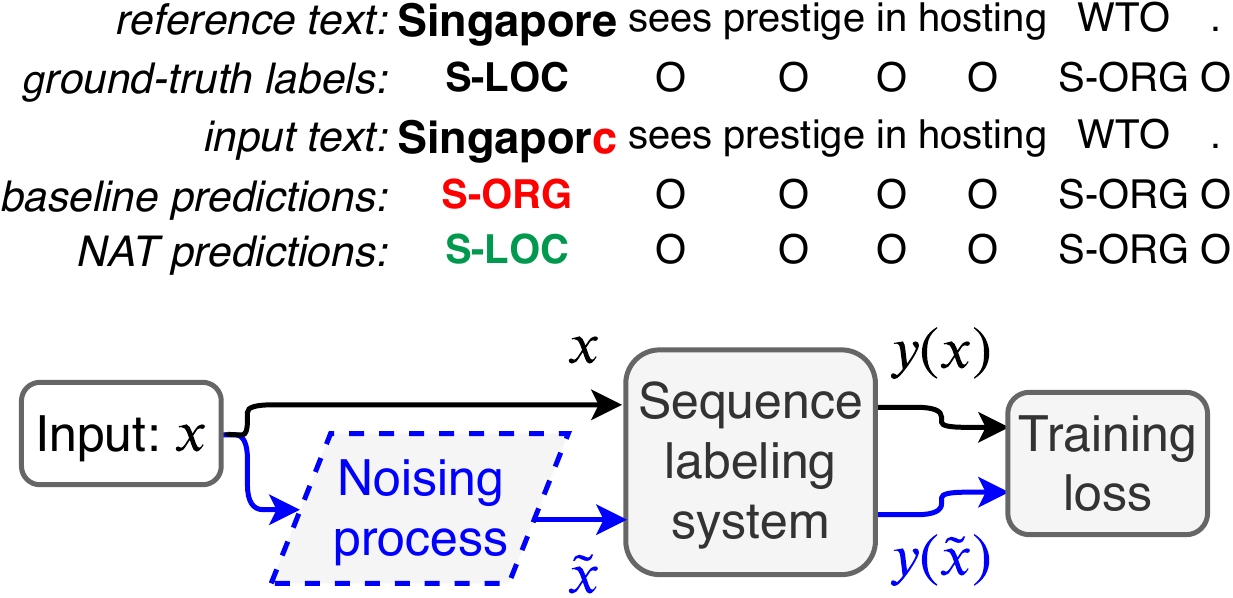}
\caption{An example of a labeling error on a  slightly perturbed sentence. Our noise-aware methods correctly predicted the location (LOC) label for the first word, as opposed to the standard approach, which misclassified it as an organization (ORG).
We complement the example with a high-level idea of our noise-aware training, where the original sentence and its noisy variant are passed together through the system. The final loss is computed based on both sets of features, which improves robustness to the input perturbations.}
\label{fig:example}
\end{center}
\end{figure}

In this work, we are concerned with the performance difference of sequence labeling performed on clean and noisy input. {\em Is it possible to narrow the gap between these two domains and design an approach that is transferable to different noise distributions at test time?}
Inspired by recent research in computer vision~\citep{DBLP:conf/cvpr/ZhengSLG16}, Neural Machine Translation (NMT;~\citealp{cheng-etal-2018-towards}), and ASR~\citep{Sperber2017b}, we propose two Noise-Aware Training (NAT) objectives that improve the accuracy of sequence labeling performed on noisy input without reducing efficiency on the original data. \Cref{fig:example} illustrates the problem and our approach.

Our contributions are as follows:
\begin{itemize}
\item We formulate a noisy sequence labeling problem, where the input undergoes an unknown noising process (\S\ref{ssec:noisy-sequence-labeling}), and we introduce a model to estimate the real error distribution (\S\ref{ssec:noise-model}). Moreover, we simulate real noisy input with a novel noise induction procedure (\S\ref{ssec:noising}).
\item We propose a \emph{data augmentation} algorithm (\S\ref{ssec:data-augmentation}) that directly induces noise in the input data to perform training of the neural model using a mixture of noisy and clean samples.
\item We implement a \emph{stability training} method \citep{DBLP:conf/cvpr/ZhengSLG16}, adapted to the sequence labeling scenario, which explicitly addresses the noisy input data problem by encouraging the model to produce a noise-invariant latent representation (\S\ref{ssec:stability-training}).
\item We evaluate our methods on real OCR \mbox{errors} and misspellings against state-of-the-art baseline models~\citep{peters-etal-2018-deep,akbik-etal-2018-contextual,devlin-etal-2019-bert} and demonstrate the effectiveness of our approach (\S\ref{sec:experiments}).
\item To support future research in this area and to make our experiments reproducible, we make our code and data publicly available\ifaclfinal\footnote{NAT repository on GitHub: \url{https://github.com/mnamysl/nat-acl2020}}\else\footnote{The code and the data were included as supplementary material and will be released online after the anonymity period.}\fi.

\end{itemize}

\section{Problem Definition}

\label{sec:the-problem}

\subsection{Neural Sequence Labeling}
\label{ssec:sequence-labeling}

\begin{figure}[!htb]
\begin{center}
\includegraphics[width=1.0\columnwidth]{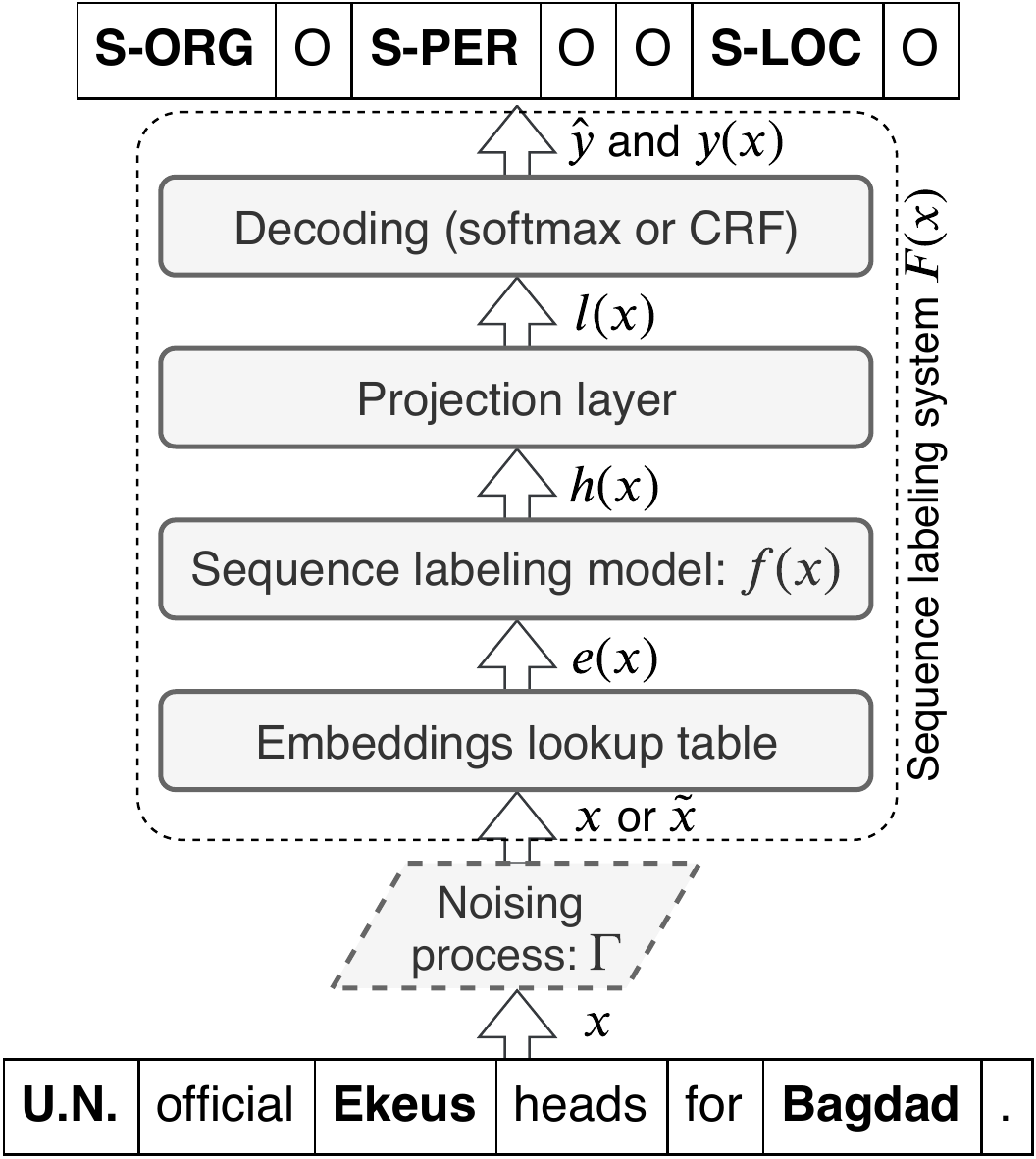}
\caption{Neural sequence labeling architecture. 
In the standard scenario (\S\ref{ssec:sequence-labeling}), the original sentence $x$ is fed as input to the sequence labeling system $F(x)$. 
Token embeddings $e(x)$ are retrieved from the corresponding look-up table and fed to the sequence labeling model $f(x)$, which outputs latent feature vectors $h(x)$.
The latent vectors are then projected to the class logits $l(x)$, which are used as input to the decoding model (softmax or CRF) that outputs the distribution over the class labels $y(x)$ and the final sequence of labels $\hat{y}$.
In a real-world scenario (\S\ref{ssec:noisy-sequence-labeling}), the input sentence undergoes an unknown noising process $\Gamma$, and the perturbed sentence $\tilde{x}$ is fed to $F(x)$. 
}
\label{fig:sequence-labeling}
\end{center}
\end{figure}

\Cref{fig:sequence-labeling} presents a typical architecture for the neural sequence labeling problem. 
We will refer to the sequence labeling system as $F(x;\theta)$, abbreviated as $F(x)$\footnote{We drop the $\theta$ parameter for brevity in the remaining of the paper. Nonetheless, we still assume that all components of $F(x;\theta)$ and all expressions derived from it also depend on $\theta$.}, where $x\myeq(x_1,\ldots,x_N)$ is a tokenized input sentence of length $N$, and $\theta$ represents all learnable parameters of the system. 
$F(x)$ takes $x$ as input and outputs the probability distribution over the class labels $y(x)$ as well as the final sequence of labels $\hat{y}\myeq(\hat{y}_1,\ldots,\hat{y}_N)$.

Either a softmax model~\citep{chiu-nichols-2016-named} or a Conditional Random Field (CRF;~\citealp{lample-etal-2016-neural}) can be used to model the output distribution over the class labels $y(x)$ from the logits $l(x)$, i.e., non-normalized predictions, and to output the final sequence of labels $\hat{y}$. 
As a labeled entity can span several consecutive tokens within a sentence, special tagging schemes are often  employed for decoding, e.g., BIOES, where the {\bf B}eginning, {\bf I}nside, {\bf O}utside, {\bf E}nd-of-entity and {\bf S}ingle-tag-entity sub-tags are also distinguished~\citep{ratinov-roth-2009-design}.
This method introduces strong dependencies between subsequent labels, which are modeled explicitly by a CRF~\citep{Lafferty:2001:CRF:645530.655813} that produces the most likely sequence of labels.

\subsection{Noisy Neural Sequence Labeling}
\label{ssec:noisy-sequence-labeling}

Similar to human readers, sequence labeling should perform reliably both in ideal and sub-optimal conditions. Unfortunately, this is rarely the case. User-generated text is a rich source of informal language containing misspellings, typos, or scrambled words~\citep{derczynski-etal-2013-twitter}. 
Noise can also be introduced in an upstream task, like OCR~\citep{Alex:2014:ERQ:2595188.2595214} or ASR~\citep{DBLP:conf/asru/ChenHHL17}, causing the errors to be propagated downstream.

To include the noise present on the source side of $F(x)$, we can modify its definition accordingly (\Cref{fig:sequence-labeling}). 
Let us assume that the input sentence $x$ is additionally subjected to some unknown noising process 
$\Gamma\myeq P(\tilde{x}_i\given{x_i})$,
where $x_i$ is the original $i$-th token, and $\tilde{x}_i$ is its distorted equivalent. 
Let $\mathcal{V}$ be the vocabulary of tokens and $\tilde{\mathcal{V}}$ be a set of all finite character sequences over an alphabet $\Sigma$.
$\Gamma$ is known as the \emph{noisy channel matrix}~\citep{brill-moore-2000-improved} and can be constructed by estimating the probability $P(\tilde{x}_i\given{x_i})$ of each distorted token $\tilde{x}_i$ given the intended token $x_i$ for every \mbox{$x_i\in\mathcal{V}$ and $\tilde{x}_i\in\tilde{\mathcal{V}}$}.

\subsection{Named Entity Recognition}
\label{ssec:the-problem}

We study the effectiveness of state-of-the-art Named Entity Recognition (NER) systems in handling imperfect input data. 
NER can be considered as a special case of the sequence labeling problem, where the goal is to locate all named entity mentions in unstructured text and to classify them into pre-defined categories, e.g., person names, organizations, and locations~\citep{tjong-kim-sang-de-meulder-2003-introduction}.
NER systems are often trained on the clean text. Consequently, they exhibit degraded performance in real-world scenarios where the transcriptions are produced by the previous upstream component, such as OCR or ASR (\S\ref{ssec:noisy-sequence-labeling}), which results in a detrimental mismatch between the training and the test conditions.
Our goal is to improve the robustness of sequence labeling performed on data from noisy sources, without deteriorating performance on the original data. 
We assume that the source sequence of tokens $x$ may contain errors. However, the noising process is generally label-preserving, i.e., the level of noise is not significant enough to affect the corresponding labels\footnote{Moreover, a human reader should be able to infer the correct label $y_i$ from the token $\tilde{x}_i$ and its context. We assume that this corresponds to a character error rate of \mbox{$\le 20\%$}.}.
It follows that the noisy token $\tilde{x}_i$ inherits the ground-truth label $y_i$ from the underlying original token $x_i$. 

\pagebreak[1]

\section{Noise-Aware Training}
\label{sec:the-method}

\subsection{Noise Model}
\label{ssec:noise-model}

To model the noise, we use the character-level noisy channel matrix $\Gamma$, which we will refer to as the \textit{character confusion matrix} (\S\ref{ssec:noisy-sequence-labeling}).

\paragraph{Natural noise}{We can estimate the natural error distribution by calculating the alignments between the pairs $(\tilde{x},x)\in\mathcal{P}$ of noisy and clean sentences using the \emph{Levenshtein distance} metric~\citep{Levenshtein1966a}, where $\mathcal{P}$ is a corpus of paired noisy and manually corrected sentences (\S\ref{ssec:noisy-sequence-labeling}). 
The allowed edit operations include insertions, deletions, and substitutions of characters. 
We can model insertions and deletions by introducing an additional symbol $\varepsilon$  into the character confusion matrix. 
The probability of insertion and deletion can then be formulated as $P_{ins}(c\given{\varepsilon})$ and $P_{del}(\varepsilon\given{c})$, where $c$ is a character to be inserted or deleted, respectively.

\paragraph{Synthetic noise}{$\mathcal{P}$ is usually laborious to obtain. Moreover, the exact modeling of noise might be impractical, and it is often difficult to accurately estimate the exact noise distribution to be encountered at test time. Such distributions may depend on, e.g., the OCR engine used to digitize the documents.
Therefore, we keep the estimated natural error distribution for evaluation and use a simplified synthetic error model for training.
We assume that all types of edit operations are equally likely:
\begin{align*}
\smashoperator[lr]{\sum_{\tilde{c}\,\in\,\Sigma{\setminus}\{\varepsilon\}}}P_{ins}(\tilde{c}\given{\varepsilon})
=
P_{del}(\varepsilon\given{c})
=
\smashoperator[lr]{\sum_{\tilde{c}\,\in\,\Sigma{\setminus}\{c,\,\varepsilon\}}}P_{subst}(\tilde{c}\given{c}),
\end{align*}
\noindent
where $c$ and $\tilde{c}$ are the original and the perturbed characters, respectively.
Moreover, $P_{ins}$ and $P_{subst}$ are uniform over the set of allowed insertion and substitution candidates, respectively.
We use the hyper-parameter $\eta$ to control the amount of noise to be induced with this method\footnote{We describe the details of our vanilla error model along with the examples of confusion matrices in the appendix.}.

\subsection{Noise Induction}
\label{ssec:noising}

Ideally, we would use the noisy sentences annotated with named entity labels for training our sequence labeling models. Unfortunately, such data is scarce.
On the other hand, labeled clean text corpora are widely available~\citep{tjong-kim-sang-de-meulder-2003-introduction,benikova-etal-2014-nosta}.
Hence, we propose to use the standard NER corpora and to induce noise into the input tokens during training synthetically.

In contrast to the image domain, which is continuous, the text domain is discrete, and we cannot directly apply continuous perturbations for written language. 
Although some works applied distortions at the level of embeddings~\citep{DBLP:conf/iclr/MiyatoDG17,yasunaga-etal-2018-robust,bekoulis-etal-2018-adversarial}, we do not have a good intuition how it changes the meaning of the underlying textual input.
Instead, we apply our noise induction procedure to generate distorted copies of the input.
For every input sentence $x$, we independently perturb each token $x_{i}\myeq(c_{1},\ldots,c_{K})$,
where $K$ is the length of $x_i$, with the following procedure (\Cref{fig:noising}):
\begin{enumerate}[label=(\arabic*)]
\item 
We insert the $\varepsilon$ symbol before the first and after every character of $x_i$ to get an extended token $x_{i}^{\prime}\myeq(\varepsilon,c_{1},\varepsilon,\ldots,\varepsilon,c_{K},\varepsilon)$. 
\item 
For every character $c^\prime_k$ of $x^\prime_i$, we sample the replacement character $\tilde{c}_k^\prime$ from the corresponding probability distribution $P(\tilde{c}_k^\prime\given{c_k^\prime})$, which can be obtained by taking a row of the character confusion matrix that corresponds to $c^\prime_k$.
As a result, we get a noisy version of the extended input token $\tilde{x}_{i}^{\prime}$.
\item We remove all $\varepsilon$ symbols from $\tilde{x}^\prime_i$ and collapse the remaining characters to obtain \mbox{a noisy token $\tilde{x}_i$}.
\end{enumerate}

\begin{figure}[!htb]
\includegraphics[width=1.0\columnwidth]{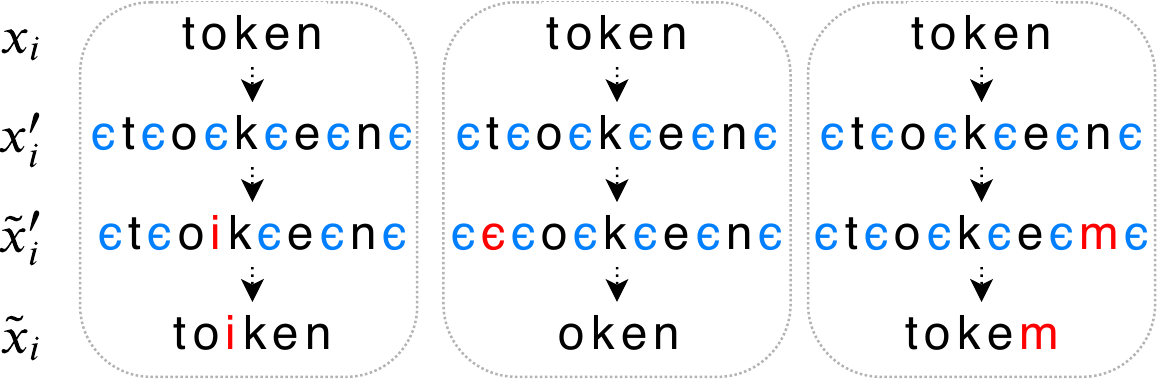}
\caption{Illustration of our noise induction procedure. Three examples correspond to insertion, deletion, and substitution errors. $x_i$, $x_{i}^{\prime}$, $\tilde{x}_{i}^{\prime}$, and $\tilde{x}_i$ are the original, extended, extended noisy, and noisy tokens, respectively.}
\label{fig:noising}
\end{figure}

\subsection{Data Augmentation Method}
\label{ssec:data-augmentation}

We can improve robustness to noise at test time by introducing various forms of artificial noise during training. We distinct regularization methods like dropout~\citep{JMLR:v15:srivastava14a} and task-specific data augmentation that transforms the data to resemble noisy input. The latter technique was successfully applied in other domains, including computer vision~\citep{Krizhevsky:2012:ICD:2999134.2999257}
and speech recognition~\citep{Sperber2017b}.

During training, we artificially induce noise into the original sentences using the algorithm described in \S\ref{ssec:noising} and train our models using a mixture of clean and noisy sentences. Let $\mathcal{L}_0(x,y;\theta)$ be the standard training objective for the sequence labeling problem, where $x$ is the input sentence, $y$ is the corresponding ground-truth sequence of labels, and $\theta$ represents the parameters of $F(x)$. We define our composite loss function as follows:
\begin{align*}
%\label{eqn:augmentation}
\begin{split}
\mathcal{L}_{augm}(x,\tilde{x},y;\theta) &\myeq \mathcal{L}_0(x,y;\theta) + \alpha\mathcal{L}_0(\tilde{x},y;\theta),
\end{split}
\end{align*}

\noindent
where $\tilde{x}$ is the perturbed sentence, and $\alpha$ is a weight of the noisy loss component. $\mathcal{L}_{augm}$ is a weighted sum of standard losses calculated using clean and noisy sentences. Intuitively, the model that would optimize $\mathcal{L}_{augm}$ should be more robust to imperfect input data, retaining the ability to perform well on clean input. \Cref{fig:data-augmentation} presents a schematic visualization of our data augmentation approach.

\subsection{Stability Training Method}
\label{ssec:stability-training}

\citet{DBLP:conf/cvpr/ZhengSLG16} pointed out the output instability issues of deep neural networks. 
They proposed a training method to stabilize deep networks against small input perturbations and applied it to the tasks of near-duplicate image detection, similar-image ranking, and image classification. 
Inspired by their idea, we adapt the stability training method to the natural language scenario. 

Our goal is to stabilize the outputs $y(x)$ of a sequence labeling system against small input perturbations, which can be thought of as flattening $y(x)$ in a close neighborhood of any input sentence $x$. 
When a perturbed copy $\tilde{x}$ is close to $x$, then $y(\tilde{x})$ should also be close to $y(x)$. 
Given the standard training objective $\mathcal{L}_0(x,y;\theta)$, the original input sentence $x$, its perturbed copy $\tilde{x}$ and the sequence of ground-truth labels $y$, we can define the stability training objective $\mathcal{L}_{stabil}$ as follows:
\begin{align*}
%\label{eqn:stability}
\begin{split}
\mathcal{L}_{stabil}(x,\tilde{x},y;\theta) 
&\myeq \mathcal{L}_0(x,y;\theta) + \alpha\mathcal{L}_{sim}(x,\tilde{x};\theta),
\\
\mathcal{L}_{sim}(x,\tilde{x};\theta) &\myeq \mathcal{D}\big(y(x), y(\tilde{x})\big),
\end{split}
\end{align*}

\noindent
where $\mathcal{L}_{sim}$ encourages the similarity of the model outputs for both $x$ and $\tilde{x}$, $\mathcal{D}$ is a task-specific feature distance measure, and $\alpha$ balances the strength of the similarity objective. 
Let $R(x)$ and $Q(\tilde{x})$ be the discrete probability distributions obtained by calculating the softmax function over the logits $l(x)$ for $x$ and $\tilde{x}$, respectively:
\begin{align*}
%\label{eqn:prob_dist}
\begin{split}
R(x) &\myeq  P(y\given{x})\myeq softmax\big(l(x)\big),
\\
Q(\tilde{x}) &\myeq  P(y\given{\tilde{x}})\myeq softmax\big(l(\tilde{x})\big).
\end{split}
\end{align*}

\noindent
We model $\mathcal{D}$ as \textit{Kullback–Leibler divergence} ($\mathcal{D}_{KL}$), which measures the correspondence between the likelihood of the original and the perturbed input:
\begin{align*}
%\label{eqn:sim}
\begin{split}
\mathcal{L}_{sim}(x,\tilde{x};\theta) &\myeq 
\sum\nolimits_{i}\mathcal{D}_{KL}\big(R(x_i)\doublevert{Q(\tilde{x}_i)}\big),
\\
\mathcal{D}_{KL}\big(R(x)\doublevert{Q(\tilde{x})}\big)
&\myeq 
\sum\nolimits_{j}P(y_{j}\given{x})\log\frac{P(y_{j}\given{x})}{P(y_{j}\given{\tilde{x}})},
\end{split}
\end{align*}

\noindent
where $i$, $j$ are the token, and the class label indices, respectively.
\Cref{fig:stability-training} summarizes the main idea of our stability training method.
\begin{figure}[!htbp]
\begin{subfigure}[!htbp]{\columnwidth}
\includegraphics[width=1.0\columnwidth]{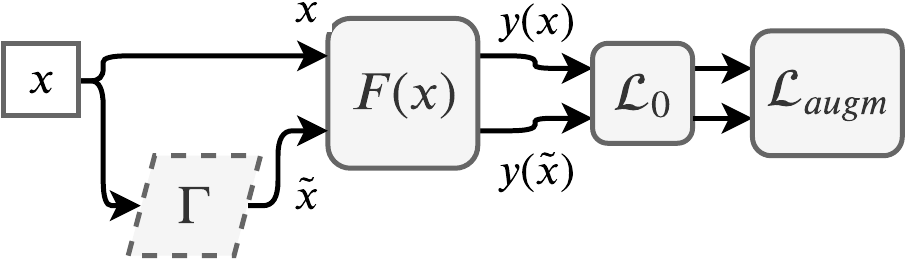}
\caption{Data augmentation training objective $\mathcal{L}_{augm}$.}
\label{fig:data-augmentation}
\end{subfigure}
%\\
%\vspace{0.1cm}
\par\smallskip % force a bit of vertical whitespace
%\\
\begin{subfigure}[!htbp]{\columnwidth}
\includegraphics[width=1.0\columnwidth]{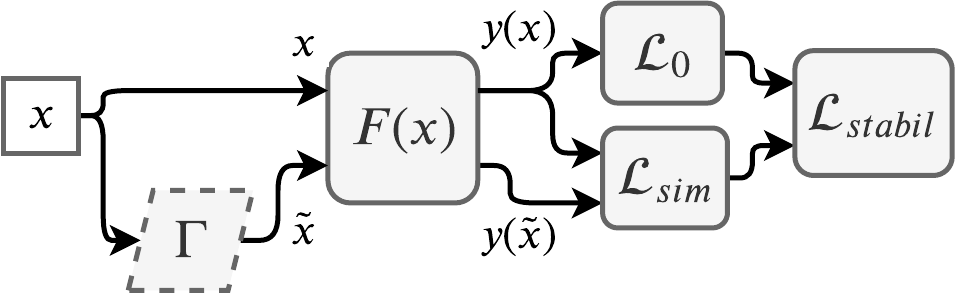}
\caption{Stability training objective $\mathcal{L}_{stabil}$.}
\label{fig:stability-training}
\end{subfigure}
\caption{Schema of our auxiliary training objectives. $x$, $\tilde{x}$ are the original and the perturbed inputs, respectively, that are fed to the sequence labeling system $F(x)$. 
$\Gamma$ represents a noising process.
$y(x)$ and $y(\tilde{x})$ are the output distributions over the entity classes for $x$ and $\tilde{x}$, respectively.
$\mathcal{L}_0$ is the standard training objective.
$\mathcal{L}_{augm}$ combines $\mathcal{L}_0$ computed on both outputs from $F(x)$.
$\mathcal{L}_{stabil}$ fuses $\mathcal{L}_0$ calculated on the original input with the similarity objective $\mathcal{L}_{sim}$.}
\label{fig:auxiliary-objectives} % Always place the \label after (or within) the caption
\end{figure}

A critical difference between the data augmentation and the stability training method is that the latter does not use noisy samples for the original task, but only for the stability objective\footnote{Both objectives could be combined and used together. However, our goal is to study their impact on robustness separately, and we leave further exploration to future work.}. Furthermore, both methods need perturbed copies of the input samples, which results in longer training time but could be  ameliorated by fine-tuning the existing model for a few epochs\footnote{We did not explore this setting in this paper, leaving such optimization to future work.}.

\section{Evaluation}
\label{sec:experiments}

\subsection{Experiment Setup}

\paragraph{Model architecture}
We used a BiLSTM-CRF architecture~\citep{DBLP:journals/corr/HuangXY15} with a single Bidirectional Long-Short Term Memory (BiLSTM) layer and $256$ hidden units in both directions for $f(x)$ in all experiments.
We considered four different text representations $e(x)$, which were used to achieve state-of-the-art results on the studied data set and should also be able to handle misspelled text and out-of-vocabulary (OOV) tokens:
\begin{itemize} %[leftmargin=*][label={--}]{$\bullet$}{$\star$}{$\circ$}{$\checkmark$}{$\square$}{$\blacksquare$}{$\bigstar$}
\item \emph{FLAIR}~\citep{akbik-etal-2018-contextual} learns a Bidirectional Language Model (BiLM) using an LSTM network to represent any sequence of characters. We used settings recommended by the authors and combined {FLAIR} with GloVe~(\citealp{pennington-etal-2014-glove}; {\it FLAIR\,+\,GloVe}) for English and Wikipedia {FastText} embeddings (\citealp{bojanowski-etal-2017-enriching};~{\it FLAIR\,+\,Wiki}) for German.
\item \emph{BERT}~\citep{devlin-etal-2019-bert} employs a Transformer encoder to learn a BiLM from large unlabeled text corpora and sub-word units to represent textual tokens. 
We use the {BERT\textsubscript{BASE}} model in our experiments.
\item \emph{ELMo}~\citep{peters-etal-2018-deep} utilizes a linear combination of hidden state vectors derived from a BiLSTM word language model trained on a large text corpus.
\item \emph{{Glove}/{Wiki}\,+\,Char} is a combination of pre-trained word embeddings ({GloVe} for English and Wikipedia {FastText} for German) and randomly initialized character embeddings~\citep{lample-etal-2016-neural}.
\end{itemize}

\paragraph{Training} 
We trained the sequence labeling model $f(x)$ and the final CRF decoding layer on top of the pre-trained embedding vectors $e(x)$, which were fixed during training, except for the character embeddings (\Cref{fig:sequence-labeling}). 
We used a mixture of the original data and its perturbed copies generated from the synthetic noise distribution (\S\ref{ssec:noise-model}) with our noise induction procedure (\S\ref{ssec:noising}).
We kept most of the hyper-parameters consistent with~\citet{akbik-etal-2018-contextual}\footnote{We list the detailed hyper-parameters in the appendix.}.
We trained our models for at most $100$ epochs and used early stopping based on the development set performance, measured as an average F1 score of clean and noisy samples.
Furthermore, we used the development sets of each benchmark data set for validation only and not for training.

\paragraph{Performance measures} We measured the entity-level micro average F1 score on the test set to compare the results of different models. 
We evaluated on both the original and the perturbed data using various natural error distributions.
We induced \textbf{OCR errors} based on the character confusion matrix $\Gamma$ (\S\ref{ssec:noising}) that was gathered on a large document corpus~\citep{8977969} using the Tesseract OCR engine~\citep{Smith2007}.
Moreover, we employed two sets of \textbf{misspellings} released by \citet{DBLP:conf/iclr/BelinkovB18} and \citet{piktus-etal-2019-misspelling}. Following the authors, we replaced every original token with the corresponding misspelled variant, sampling uniformly among available replacement candidates.
We present the estimated error rates of text that is produced with these noise induction procedures \notinsubfile{in \Cref{tab:error-rates}} in the appendix.
As the evaluation with noisy data leads to some variance in the final scores, we repeated all experiments five times and reported mean and standard deviation. 

\paragraph{Implementation} We implemented our models using the {FLAIR} framework~\citep{akbik-etal-2019-flair}\footnote{We used {FLAIR} v0.4.2.}.
We extended their sequence labeling model by integrating our auxiliary training objectives (\S\ref{ssec:data-augmentation}, \S\ref{ssec:stability-training}). Nonetheless, our approach is universal and can be implemented in any other sequence labeling framework.

\subsection{Sequence Labeling on Noisy Data}
\label{ssec:eval1}

To validate our approach, we trained the baseline models with and without our auxiliary loss objectives (\S\ref{ssec:data-augmentation},~\S\ref{ssec:stability-training})\footnote{We experimented with a pre-processing step that used a spell checking module, but it did not provide any benefits and even decreased accuracy on the original data. Therefore we did not consider it a viable solution for this problem.}. 
We used the CoNLL 2003~\citep{tjong-kim-sang-de-meulder-2003-introduction} and the GermEval 2014~\citep{benikova-etal-2014-nosta} data sets in this setup\footnote{We present data set statistics and sample outputs from our system in the appendix.}.
The baselines utilized GloVe vectors coupled with FLAIR and character embeddings ({FLAIR\,+\,GloVe}, {GloVe\,+\,Char}), {BERT}, and {ELMo} embeddings for English. For German, we employed Wikipedia FastText vectors paired with FLAIR and character embeddings ({FLAIR\,+\,Wiki}, {Wiki\,+\,Char})\footnote{This choice was motivated by the availability of pre-trained embedding models in the {FLAIR} framework.}. 
We used a label-preserving training setup ($\alpha\myeq1.0$, $\eta_{train}\myeq10\%$).

{\renewcommand{\arraystretch}{1.0}\setlength{\tabcolsep}{1.0pt}
\begin{table*}[!htbp]
\centering \small 
\begin{tabular}{C{0.09}C{0.08}L{0.1}L{0.13}L{0.19}L{0.19}L{0.19}}
\toprule
Data set & Model & Train loss & Original data & OCR errors & Misspellings$^\dag$ & Misspellings$^\ddag$  \\
\midrule
\multirow{12}{0.08\textwidth}[-12pt]{\centering English CoNLL 2003} & \multirow{3}{0.08\textwidth}{\centering{FLAIR + GloVe}} 
& $\mathcal{L}_0$ & 92.05 & \meanstddev{76.44}{0.45} & \meanstddev{75.09}{0.48} & \meanstddev{87.57}{0.10}\\
& & $\mathcal{L}_{augm}$ & {\bf 92.56} \posdiff{0.51} & {\bf\meanstddev{84.79}{0.23}} \posdiff{8.35} & {\bf\meanstddev{83.57}{0.43}} \posdiff{8.48} & {\bf\meanstddev{90.50}{0.08}} \posdiff{2.93}\\
& & $\mathcal{L}_{stabil}$ & 91.99 \negdiff{0.06} & \meanstddev{84.39}{0.37} \posdiff{7.95} & \meanstddev{82.43}{0.23} \posdiff{7.34} & \meanstddev{90.19}{0.14} \posdiff{2.62}\\
\cmidrule{2-7}
& \multirow{3}{*}{{BERT}}
& $\mathcal{L}_0$ & 90.91 & \meanstddev{68.23}{0.39} & \meanstddev{65.65}{0.31} & \meanstddev{85.07}{0.15}\\
& & $\mathcal{L}_{augm}$ & 90.84 \negdiff{0.07} & {\bf\meanstddev{79.34}{0.32}} \posdiff{11.11} & {\bf\meanstddev{75.44}{0.28}} \posdiff{9.79} & \meanstddev{86.21}{0.24} \posdiff{1.14}\\
& & $\mathcal{L}_{stabil}$ & {\bf 90.95} \posdiff{0.04} & \meanstddev{78.22}{0.17} \posdiff{9.99} & \meanstddev{73.46}{0.34} \posdiff{7.81} & {\bf\meanstddev{86.52}{0.12}} \posdiff{1.45}\\
\cmidrule{2-7}
& \multirow{3}{*}{{ELMo}}
& $\mathcal{L}_0$ & {\bf 92.16} & \meanstddev{72.90}{0.50} & \meanstddev{70.99}{0.17} & \meanstddev{88.59}{0.19}\\
& & $\mathcal{L}_{augm}$ & 91.85 \negdiff{0.31} & {\bf\meanstddev{84.09}{0.18}} \posdiff{11.19} & {\bf\meanstddev{82.33}{0.40}} \posdiff{11.34} & {\bf\meanstddev{89.50}{0.16}} \posdiff{0.91}\\
& & $\mathcal{L}_{stabil}$ & 91.78 \negdiff{0.38} & \meanstddev{83.86}{0.11} \posdiff{10.96} & \meanstddev{81.47}{0.29} \posdiff{10.48} & {\bf\meanstddev{89.49}{0.15}} \posdiff{0.90}\\
\cmidrule{2-7}
& \multirow{3}{0.08\textwidth}{\centering{GloVe + Char}}
& $\mathcal{L}_0$ & 90.26 & \meanstddev{71.15}{0.51} & \meanstddev{70.91}{0.39} & \meanstddev{87.14}{0.07}\\
& & $\mathcal{L}_{augm}$ & {\bf 90.83} \posdiff{0.57} & {\bf \meanstddev{81.09}{0.47}} \posdiff{9.94} & {\bf\meanstddev{79.47}{0.24}} \posdiff{8.56} & {\bf\meanstddev{88.82}{0.06}} \posdiff{1.68}\\
& & $\mathcal{L}_{stabil}$ & 90.21 \negdiff{0.05} & \meanstddev{80.33}{0.29} \posdiff{9.18} & \meanstddev{78.07}{0.23} \posdiff{7.16} & \meanstddev{88.47}{0.13} \posdiff{1.33}\\
\midrule
\multirow{6}{0.08\textwidth}[-5pt]{\centering German CoNLL 2003} & \multirow{3}{0.08\textwidth}{\centering{FLAIR + Wiki}}
& $\mathcal{L}_0$ & 86.13 & \meanstddev{66.93}{0.49} & \meanstddev{78.06}{0.13} & \meanstddev{80.72}{0.23}  \\
& & $\mathcal{L}_{augm}$ & {\bf 86.46} \posdiff{0.33} & {\bf\meanstddev{75.90}{0.63}} \posdiff{8.97} & {\bf\meanstddev{83.23}{0.14}} \posdiff{5.17} & \meanstddev{84.01}{0.27} \posdiff{3.29}\\
& & $\mathcal{L}_{stabil}$ & 86.33 \posdiff{0.20} & \meanstddev{75.08}{0.29} \posdiff{8.15} & \meanstddev{82.60}{0.21} \posdiff{4.54} & {\bf\meanstddev{84.12}{0.26}} \posdiff{3.40} \\
\cmidrule{2-7}
& \multirow{3}{0.08\textwidth}{\centering{Wiki + Char}}
& $\mathcal{L}_0$ & 82.20 & \meanstddev{59.15}{0.76} & \meanstddev{75.27}{0.31} & \meanstddev{71.45}{0.15} \\
& & $\mathcal{L}_{augm}$ & {\bf 82.62} \posdiff{0.42} & \meanstddev{67.67}{0.75} \posdiff{8.52} & {\bf\meanstddev{78.48}{0.24}} \posdiff{3.21} & \meanstddev{79.14}{0.31} \posdiff{7.69}\\
& & $\mathcal{L}_{stabil}$ & 82.18 \negdiff{0.02} & {\bf\meanstddev{67.72}{0.63}} \posdiff{8.57} & \meanstddev{77.59}{0.12} \posdiff{2.32} & {\bf\meanstddev{79.33}{0.39}} \posdiff{7.88}\\
\midrule
\multirow{6}{0.08\textwidth}[-5pt]{\centering Germ-Eval 2014} & \multirow{3}{0.08\textwidth}{\centering{FLAIR + Wiki}}
& $\mathcal{L}_0$ & {\bf 85.05} & \meanstddev{58.64}{0.51} & \meanstddev{67.96}{0.23} & \meanstddev{68.64}{0.28} \\
& & $\mathcal{L}_{augm}$ & 84.84 \negdiff{0.21} & {\bf\meanstddev{72.02}{0.24}} \posdiff{13.38} & {\bf\meanstddev{78.59}{0.11}} \posdiff{10.63} & {\bf\meanstddev{81.55}{0.12}} \posdiff{12.91}\\
& & $\mathcal{L}_{stabil}$ & 84.43 \negdiff{0.62} & \meanstddev{70.15}{0.27} \posdiff{11.51} & \meanstddev{75.67}{0.16} \posdiff{7.71} & \meanstddev{79.31}{0.32} \posdiff{10.67}\\
\cmidrule{2-7}
& \multirow{3}{0.08\textwidth}{\centering{Wiki + Char}}
& $\mathcal{L}_0$ & 80.32 & \meanstddev{52.48}{0.31} & \meanstddev{61.99}{0.35} & \meanstddev{54.86}{0.15} \\
& & $\mathcal{L}_{augm}$ & {\bf 80.68} \posdiff{0.36} & {\bf \meanstddev{63.74}{0.31}} \posdiff{11.26} & {\bf\meanstddev{70.83}{0.09}} \posdiff{8.84} & {\bf\meanstddev{75.66}{0.11}} \posdiff{20.80}\\
& & $\mathcal{L}_{stabil}$ & 80.00 \negdiff{0.32} & \meanstddev{62.29}{0.35} \posdiff{9.81} & \meanstddev{68.23}{0.23} \posdiff{6.24} & \meanstddev{72.40}{0.29} \posdiff{17.54}\\
\bottomrule
\end{tabular}
\caption{Evaluation results on the CoNLL 2003 and the GermEval 2014 test sets. We report results on the original data, as well as on its noisy copies with OCR errors and two types of misspellings released by \citet{DBLP:conf/iclr/BelinkovB18}$^\dag$ and \citet{piktus-etal-2019-misspelling}$^\ddag$. 
$\mathcal{L}_0$ is the standard training objective. 
$\mathcal{L}_{augm}$ and $\mathcal{L}_{stabil}$ are the data augmentation and the stability objectives, respectively. 
We report mean F1 scores with standard deviations from five experiments 
and mean differences against the standard objective (in parentheses). 
}
\label{tab:eval1}
\end{table*}}

\Cref{tab:eval1} presents the results of this experiment\footnote{We did not replicate the exact results from the original papers because we did not use development sets for training, and our approach is feature-based, as we did not fine-tune embeddings on the target task.
}.
We found that our auxiliary training objectives boosted accuracy on noisy input data for all baseline models and both languages.
At the same time, they preserved accuracy for the original input.
The data augmentation objective seemed to perform slightly better than the stability objective. However, the chosen hyper-parameter values were rather arbitrary, as our goal was to prove the utility and the flexibility of both objectives.

\subsection{Sensitivity Analysis}
\label{ssec:eval2}

We evaluated the impact of our hyper-parameters on the sequence labeling accuracy using the English CoNLL 2003 data set. 
We trained multiple models with different amounts of noise $\eta_{train}$ and different weighting factors $\alpha$.
We chose the {FLAIR\,+\,GloVe} model as our baseline because it achieved the best results in the preliminary analysis (\S\ref{ssec:eval1}) and showed good performance, which enabled us to perform extensive experiments.

\begin{figure*}[!htbp]
\centering
\begin{subfigure}[t]{0.49\textwidth}
\centering
\includegraphics[height=3.6cm]{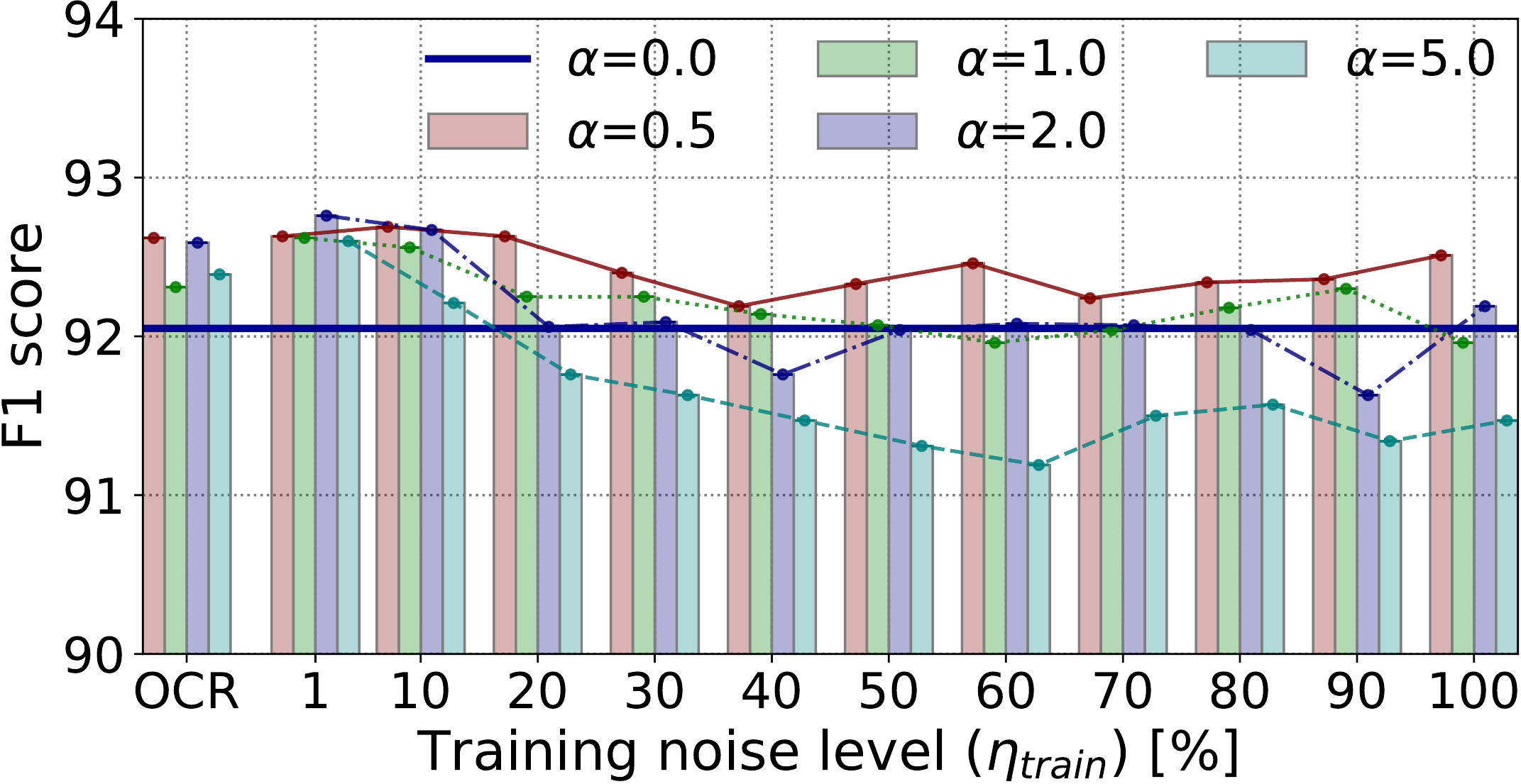}
\caption{Data augmentation objective (original test data).}
\end{subfigure}
%~ 
\begin{subfigure}[t]{0.49\textwidth}
\centering
\includegraphics[height=3.6cm]{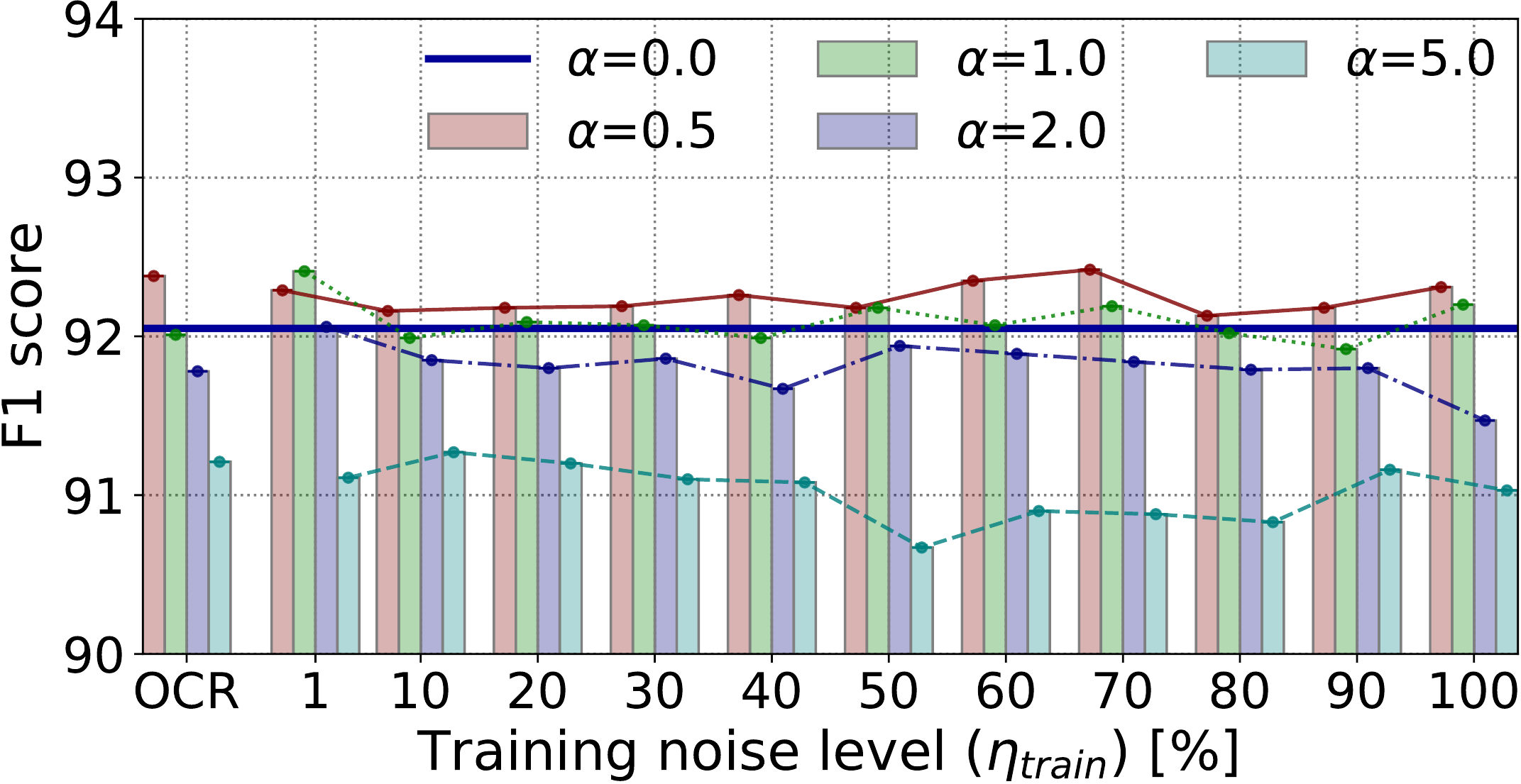}
\caption{Stability training objective (original test data).}
\end{subfigure}
%~
%\vspace{2mm}
\par\smallskip % force a bit of vertical whitespace
%\\
\begin{subfigure}[t]{0.49\textwidth}
\centering
\includegraphics[height=3.6cm]{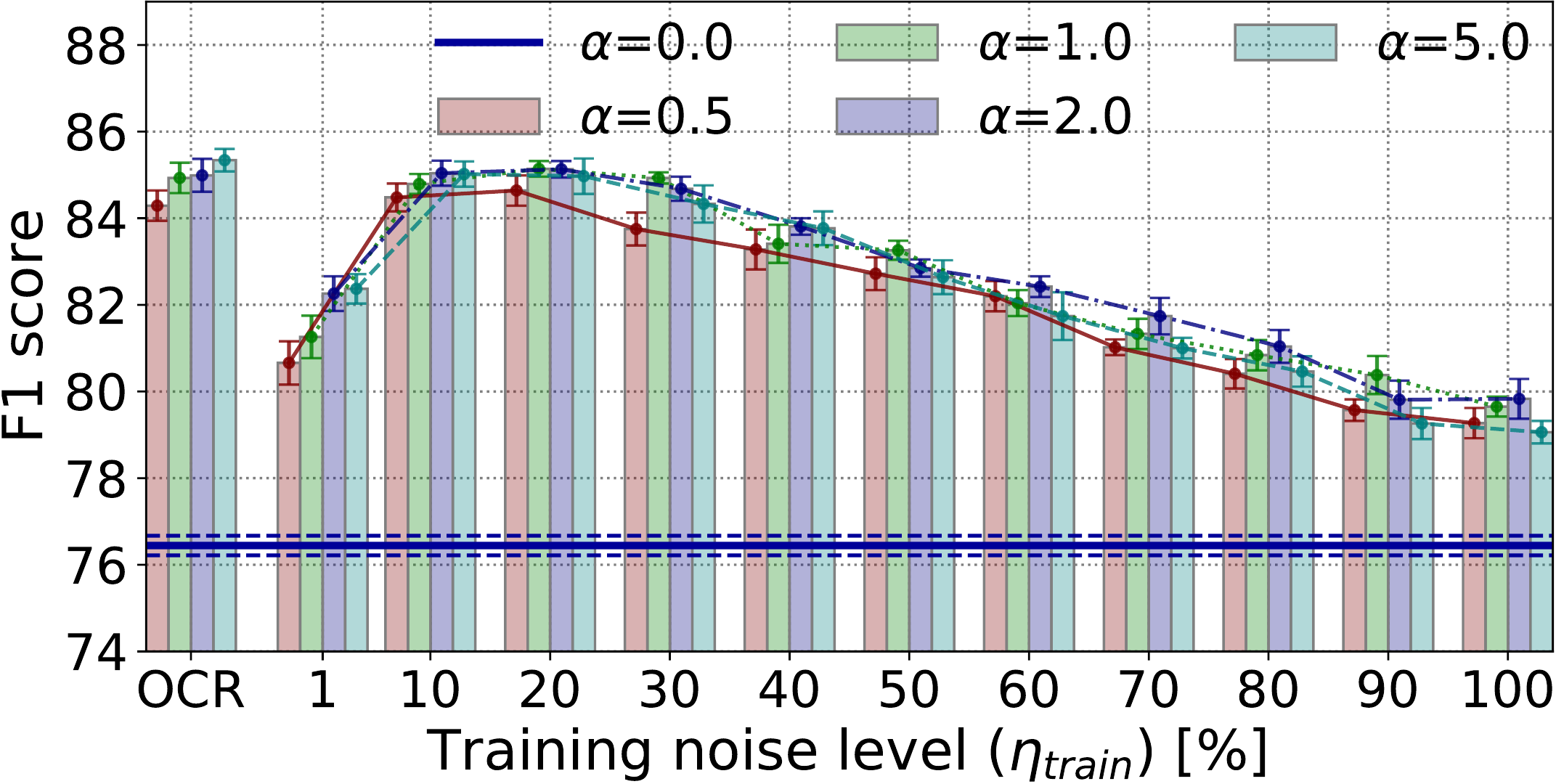}
\caption{Data augmentation objective (tested on OCR errors)}
\end{subfigure}
%~ 
\begin{subfigure}[t]{0.49\textwidth}
\centering
\includegraphics[height=3.6cm]{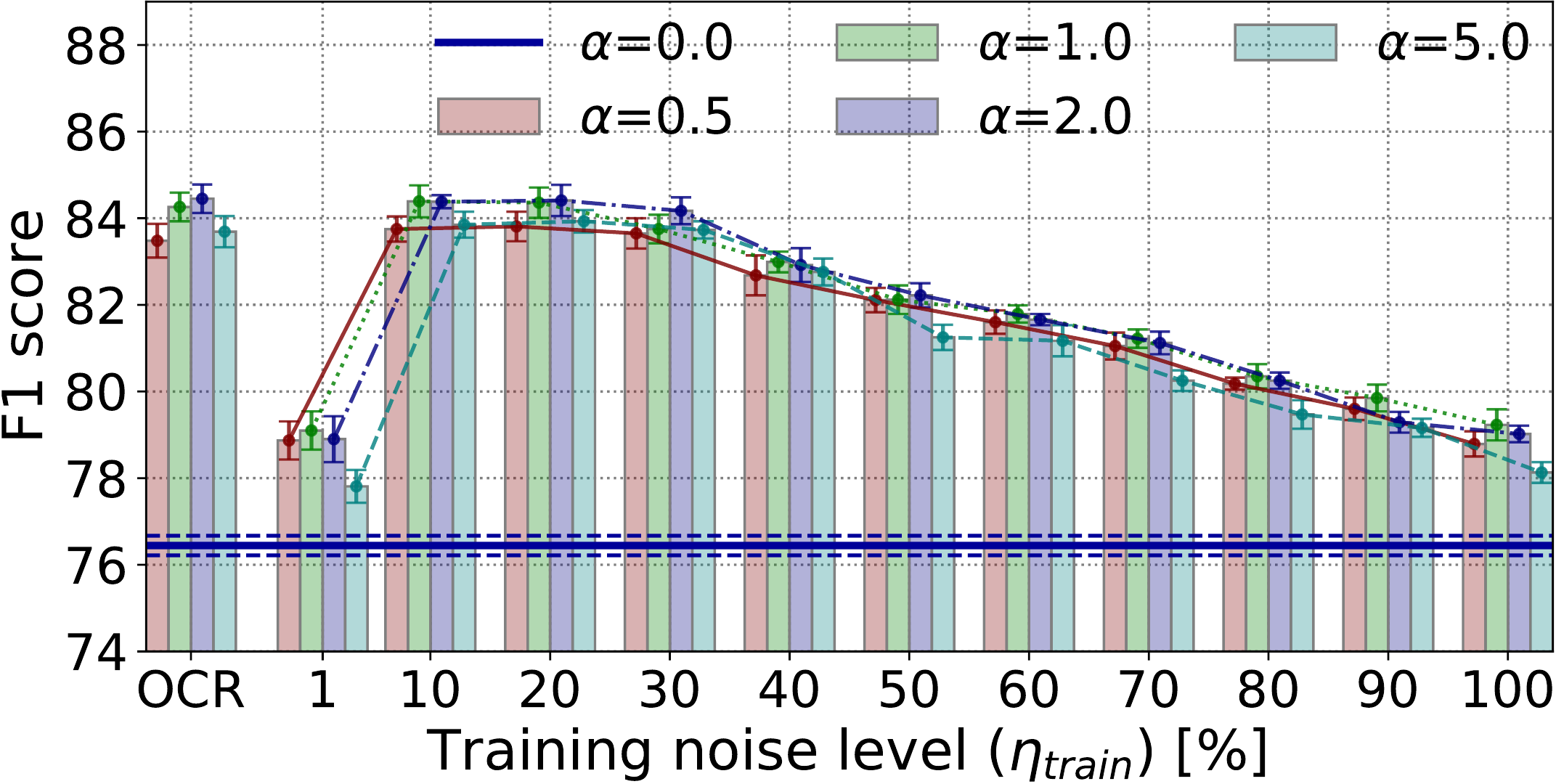}
\caption{Stability training objective (tested on OCR errors)}
\end{subfigure}         
\caption{Sensitivity analysis performed on the English CoNLL 2003 test set (\S\ref{ssec:eval2}). Each figure presents the results of models trained using one of our auxiliary training objectives on either original data or its variant perturbed with OCR errors. The bar marked as "OCR" represents a model trained using the OCR noise distribution. Other bars correspond to models trained using synthetic noise distribution and different hyper-parameters ($\alpha$, $\eta_{train}$).}
\label{fig:eval2}
\end{figure*}

\Cref{fig:eval2} summarizes the results of the sensitivity experiment. 
The models trained with our auxiliary objectives mostly preserved or even improved accuracy on the original data compared to the baseline model ($\alpha\myeq0$).
Moreover, they significantly outperformed the baseline on data perturbed with natural noise.
The best accuracy was achieved for $\eta_{train}$ from $10$ to $30\%$, which roughly corresponds to the label-preserving noise range.
Similar to \citet{heigold-etal-2018-robust} and \citet{cheng-etal-2019-robust}, we conclude that a non-zero noise level induced during training $({\eta_{train}>0})$ always yields improvements on noisy input data when compared with the models trained exclusively on clean data.
The best choice of $\alpha$ was in the range from $0.5$ to $2.0$. $\alpha=5.0$ exhibited lower performance on the original data.
Moreover, the models trained on the real error distribution demonstrated at most slightly better performance, which indicates that the exact noise distribution does not \mbox{necessarily have to be known at training time}\footnote{Nevertheless, the aspect of mimicking an empirical noise distribution requires more thoughtful analysis, and therefore we leave to future work.}.

\subsection{Error Analysis}
\label{sec:error-analysis}

To quantify improvements provided by our approach, we measured sequence labeling accuracy on the subsets of data with different levels of perturbation, i.e., we divided input tokens based on edit distance to their clean counterparts. Moreover, we partitioned the data by named entity class to assess the impact of noise on recognition of different entity types.
For this experiment, we used both the test and the development parts of the English CoNLL 2003 data set and induced OCR errors with our noising procedure.

\Cref{fig:error_analysis} presents the results for the baseline and the proposed methods.
It can be seen that our approach achieved significant error reduction across all perturbation levels and all entity types.
Moreover, by narrowing down the analysis to perturbed tokens, we discovered that the baseline model was particularly sensitive to noisy tokens from the LOC and the MISC categories. 
Our approach considerably reduced this negative effect.
Furthermore, as the stability training worked slightly better on the LOC class and the data augmentation was more accurate on the ORG type, we argue that both methods could be combined to enhance overall sequence labeling accuracy further.
Note that even if the particular token was not perturbed, its context could be noisy, which would explain the fact that our approach provided improvements even for tokens without perturbations.

\begin{figure}[htb]
\begin{subfigure}[t]{0.95\columnwidth}
\centering
\includegraphics[width=0.95\columnwidth]{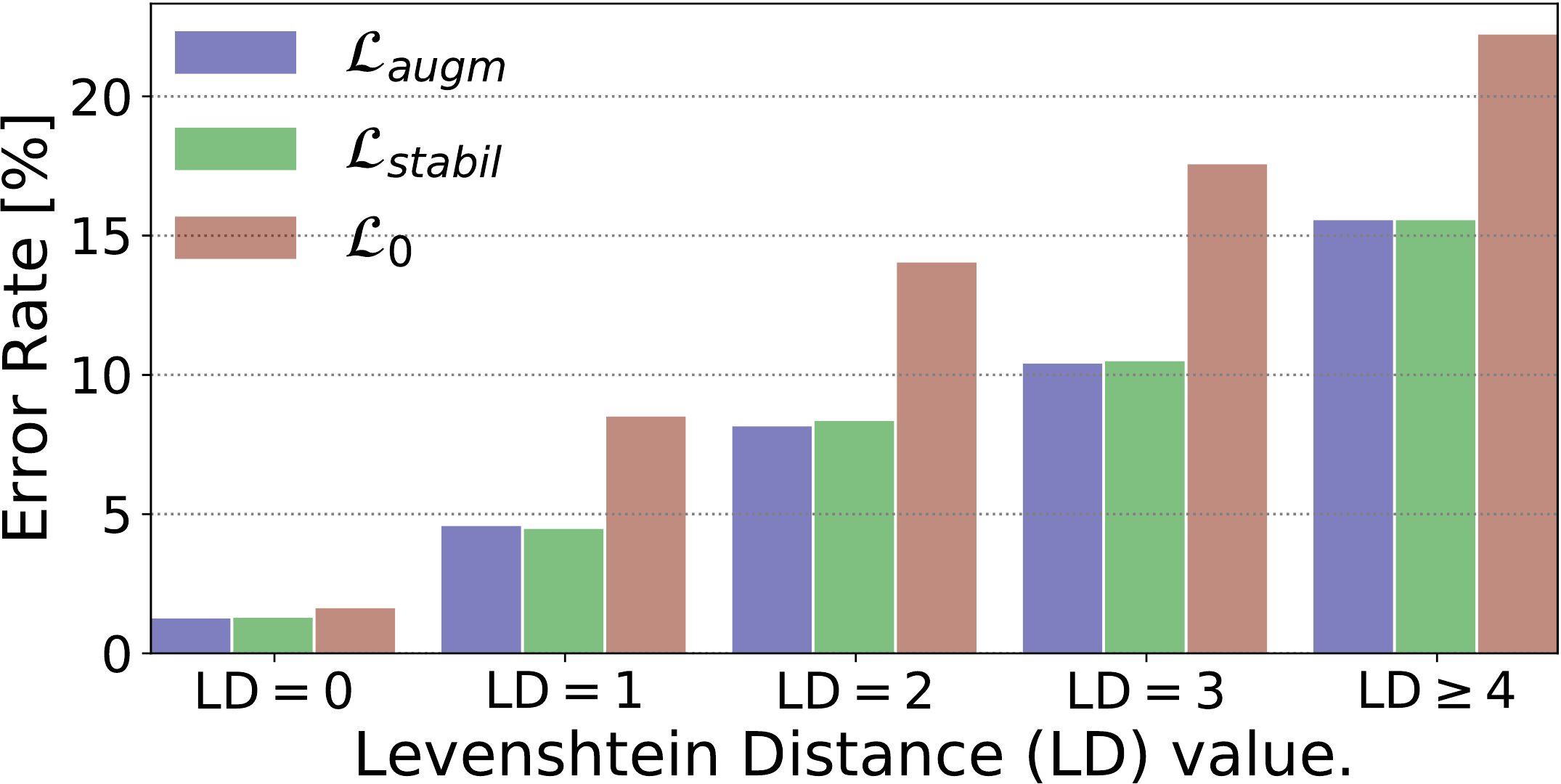}
\caption{Divided by the edit distance value.}
\label{fig:error_analysis-ld}
\end{subfigure}
\par\smallskip % force a bit of vertical whitespace
\begin{subfigure}[t]{0.95\columnwidth}
\centering
\includegraphics[width=0.95\columnwidth]{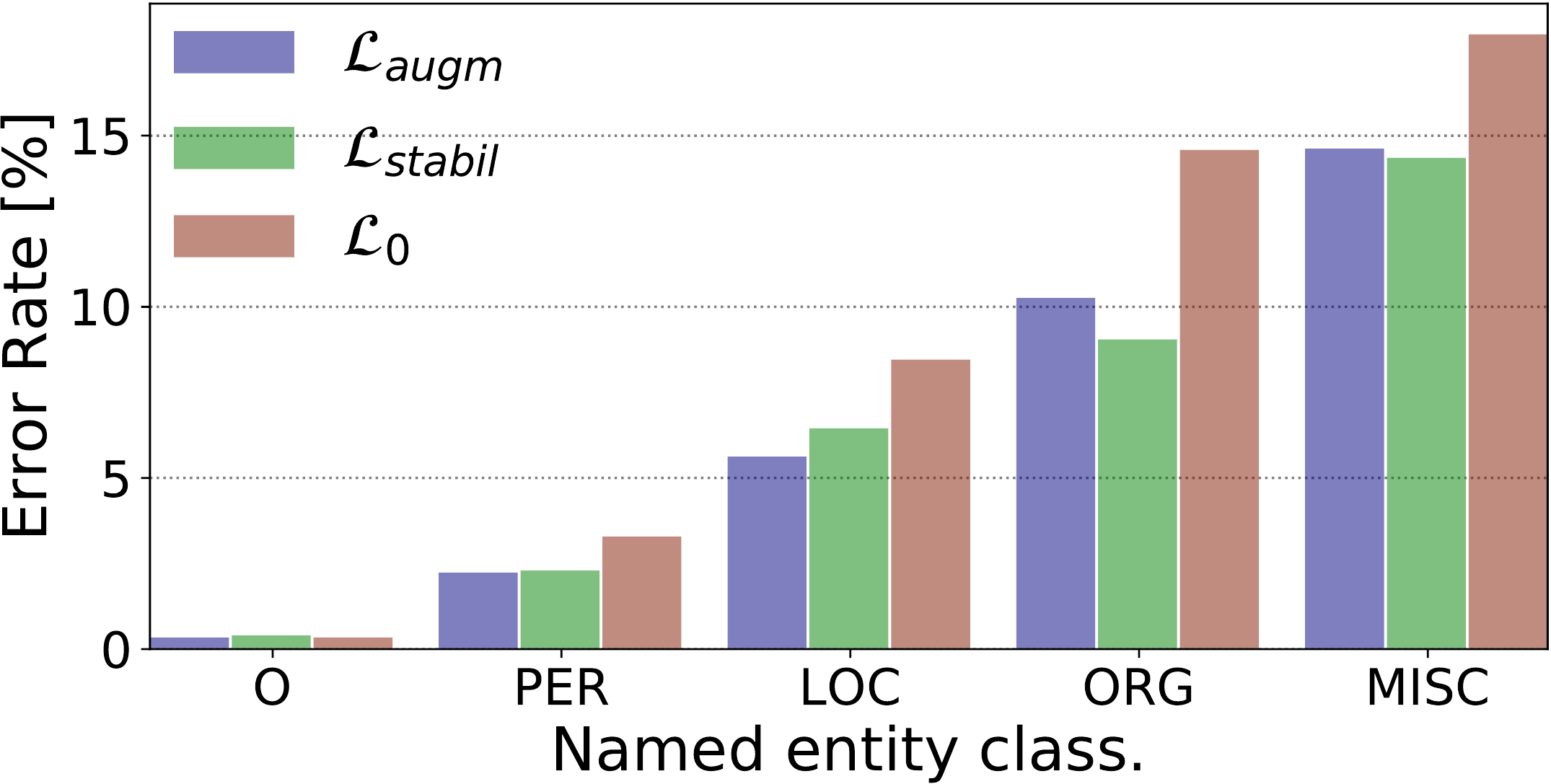}
\caption{Divided by the entity class (clean tokens).}
\label{fig:error_analysis-tags-clean}
\end{subfigure}
\par\smallskip % force a bit of vertical whitespace
\begin{subfigure}[t]{0.97\columnwidth}
\centering
\includegraphics[width=0.97\columnwidth]{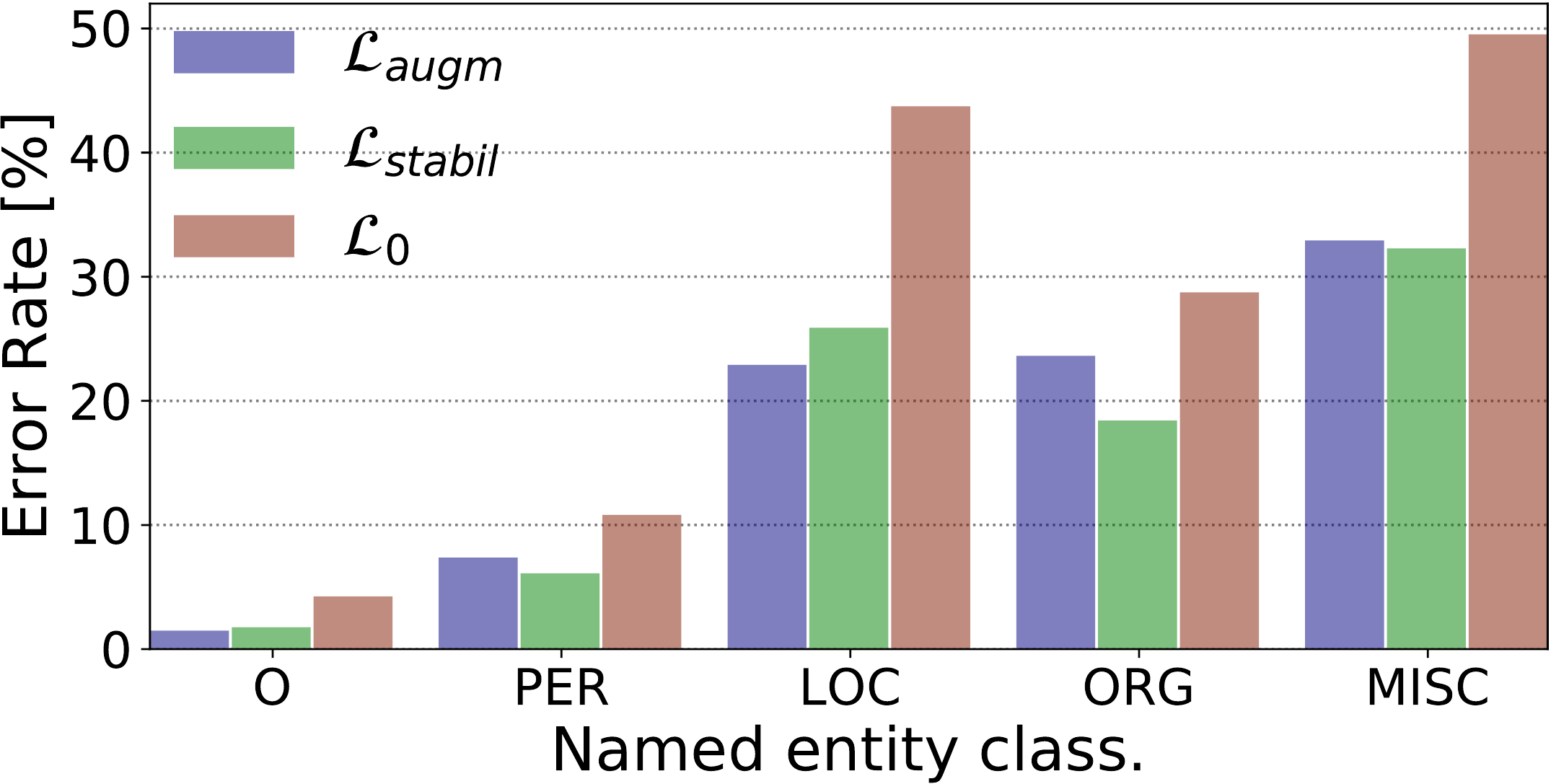}
\caption{Divided by the entity class (perturbed tokens).}
\label{fig:error_analysis-tags-noisy}
\end{subfigure}
\caption{Error analysis results on the English CoNLL 2003 data set with OCR noise.
We presented the results of the {FLAIR\,+\,GloVe} model trained with the standard and the proposed objectives.
The data was divided into the subsets based on the edit distance of a token to its original counterpart and its named entity class. The latter group was further partitioned into the clean and the perturbed tokens. 
The error rate is the percentage of tokens with misrecognized entity class labels.
}
\label{fig:error_analysis}
\end{figure}

\makeatletter
\let\maintitle\@title
\title{
\maintitle\\
(Related Work)
}
\makeatother

\maketitle

\section{Related Work}
\label{sec:related-work}

Improving robustness has been receiving increasing attention in the NLP community.
The most relevant research was conducted in the NMT domain.

\paragraph{Noise-additive data augmentation}

A natural strategy to improve robustness to noise is to augment the training data with samples perturbed using a similar noise model.
\citet{heigold-etal-2018-robust} demonstrated that the noisy input substantially degrades the accuracy of models trained on clean data. 
They used word scrambling, as well as character flips and swaps as their noise model, and achieved the best results under matched training and test noise conditions. 
\citet{DBLP:conf/iclr/BelinkovB18} reported significant degradation in the performance of NMT systems on noisy input. They built a look-up table of possible lexical replacements from Wikipedia edit histories and used it as a natural source of the noise. Robustness to noise was only achieved by training with the same distribution---at the expense of performance degradation on other types of noise. In contrast, our method performed well on natural noise at test time by using a simplified synthetic noise model during training.
\citet{karpukhin-etal-2019-training} pointed out that existing NMT approaches are very sensitive to spelling mistakes and proposed to augment training samples with random character deletions, insertions, substitutions, and swaps. 
They showed improved robustness to natural noise, represented by frequent corrections in Wikipedia edit logs, without diminishing performance on the original data.
However, not every word in the vocabulary has a corresponding misspelling.
Therefore, even when noise is applied at the maximum rate, only a subset of tokens is perturbed (20-50\%, depending on the language).
In contrast, we used a confusion matrix, which is better suited to model statistical error distribution and can be applied to all tokens, not only those present in the corresponding look-up tables. 

\paragraph{Robust representations} Another method to improve robustness is to design a representation that is less sensitive to noisy input.
\citet{DBLP:conf/cvpr/ZhengSLG16} presented a general method to stabilize model predictions against small input distortions.
\citet{cheng-etal-2018-towards} continued their work and developed the adversarial stability training method for NMT by adding a discriminator term to the objective function.
They combined data augmentation and stability objectives, while we evaluated both methods separately and provided evaluation results on natural noise distribution.
\citet{piktus-etal-2019-misspelling} learned representation that embeds misspelled words close to their correct variants. Their Misspelling Oblivious Embeddings (MOE) model jointly optimizes two loss functions, each of which iterates over a separate data set (a corpus of text and a set of misspelling/correction pairs) during training. In contrast, our method does not depend on any additional resources and uses a simplified error distribution during training.

\paragraph{Adversarial learning}

{\it Adversarial attacks} seek to mislead the neural models by feeding them with adversarial examples~\cite{42503}. 
In a white-box attack scenario~\cite{43405,ebrahimi-etal-2018-hotflip} we assume that the attacker has access to the model parameters, in contrast to the black-box scenario~\cite{alzantot-etal-2018-generating,8424632}, where the attacker can only sample model predictions on given examples. 
{\it Adversarial training}~\cite{DBLP:conf/iclr/MiyatoDG17,yasunaga-etal-2018-robust}, on the other hand, aims to improve the robustness of the neural models by utilizing adversarial examples during training.

\paragraph{The impact of noisy input data} 

In the context of ASR, 
\citet{DBLP:conf/interspeech/ParadaDJ11} observed that named entities are often OOV tokens, and therefore they cause more recognition errors.
In the document processing field, \citet{Alex:2014:ERQ:2595188.2595214} studied NER performed on several digitized historical text collections and showed that OCR errors have a significant impact on the accuracy of the downstream task.
\citet{8977969} examined the efficiency of modern OCR engines and showed that although the OCR technology was more advanced than several years ago when many historical archives were digitized~\cite{kim-cassidy-2015-finding,neudecker-2016-open}, the most widely used engines still had difficulties with non-standard or lower quality input.

\paragraph{Spelling- and post-OCR correction.}

A natural method of handling erroneous text is to correct it before feeding it to the downstream task.  
Most popular post-correction techniques include 
correction candidates ranking~\citep{fivez-etal-2017-unsupervised,flor-etal-2019-benchmark},
noisy channel modeling~\citep{brill-moore-2000-improved,duan2011online},
voting~\citep{6628604},
sequence to sequence models~\cite{afli-etal-2016-using,schmaltz-etal-2017-adapting} 
and hybrid systems~\cite{schulz-kuhn-2017-multi}.

In this paper, we have taken a different approach and attempted to make our models robust without relying on prior error correction, which, in case of OCR errors, is still far from being solved~\cite{8270163,8978127}. 

\onlyinsubfile{
\bibliographystyle{acl_natbib}
\bibliography{anthology,paper-acl2020}
}

\section{Conclusions}
\label{sec:conclusions}

In this paper, we investigated the difference in accuracy between sequence labeling performed on clean and noisy text (\S\ref{ssec:the-problem}). 
We formulated the noisy sequence labeling problem (\S\ref{ssec:noisy-sequence-labeling})
and introduced a model that can be used to estimate the real noise distribution (\S\ref{ssec:noise-model}). We developed the noise induction procedure that simulates the real noisy input (\S\ref{ssec:noising}). We proposed two noise-aware training methods that boost sequence labeling accuracy on the perturbed text:
\begin{enumerate*}[label=(\roman*)] %\arabic\alph\roman
\item Our data augmentation approach uses a mixture of clean and noisy examples during training to make the model resistant to erroneous input (\S\ref{ssec:data-augmentation}).
\item Our stability training algorithm encourages output similarity for the original and the perturbed input, which helps the model to build a noise invariant latent representation (\S\ref{ssec:stability-training}).
\end{enumerate*}
Our experiments confirmed that NAT consistently improved efficiency of popular sequence labeling models on data perturbed with different error distributions, preserving accuracy on the original input (\S\ref{sec:experiments}).
Moreover, we avoided expensive re-training of embeddings on noisy data sources by employing existing text representations.
We conclude that NAT makes existing models applicable beyond the idealized scenarios.
It may support an automatic correction method that uses recognized entity types to narrow the list of feasible correction candidates.
Another application is data anonymization~\citep{7743936}.

Future work will involve improvements in the proposed noise model to study the importance of fidelity to real-world error patterns.
Moreover, we plan to evaluate NAT on other real noise distributions (e.g., from ASR) and other sequence labeling tasks to support our claims further.

% ========= ACKNOWLEDGMENT ===========

\section*{Acknowledgments}

We would like to thank the reviewers for the time they invested in evaluating our paper and for their insightful remarks and valuable suggestions.

% =========== REFERENCES ============
% * References (1-2 pages)

\bibliographystyle{acl_natbib}
\bibliography{anthology,paper-acl2020}

\pagebreak[1]

\ifaclfinal 

\makeatletter
\let\maintitle\@title
\title{
\maintitle\\
%(Supplementary Material)
(Appendices)
}
\makeatother

\maketitle

\appendix

% ====================== Noise Model - Supplementary Materials ======================

\section{Noise Model - Supplementary Materials}
\label{sec:eval-quantitative}

In this section, we present the extended description of our vanilla noise model \notinsubfile{ introduced in \S\ref{ssec:noise-model}}.
Let $P_{edit} \myeq \eta/{3}$ be the probability of performing a single character edit operation (insertion, deletion, or substitution) that replaces the source character $c$ with a noisy character $\tilde{c}$, where $\tilde{c}\myneq{c}$.
\Cref{eq:vanilla-model} 
%\clearpage\noindent
defines the vanilla error distribution, which we use at training time:

\begin{subequations}
\label{eq:vanilla-model}
\begin{empheq}[left={P(\tilde{c}\given{c})\myeq\empheqlbrace}]{align}
\frac{P_{edit}}{\norm{\Sigma{\setminus}\{\varepsilon\}}}, & \mkern10mu\text{if}\ c\myeq\varepsilon\ \text{and}\ \tilde{c}\myneq{\varepsilon}.\label{eq:vanilla-model-ins}\\ % INSERTION
1-P_{edit}, & \mkern10mu\text{if}\ c\myeq\varepsilon\ \text{and}\ \tilde{c}\myeq{\varepsilon}.\label{eq:vanilla-model-ins-no-change}\\ % NO-CHANGE (e->e)
\frac{P_{edit}}{\norm{\Sigma{\setminus}\{c,\varepsilon\}}}, & \mkern10mu\text{if}\ c\myneq\varepsilon\ \text{and}\ \tilde{c}\myneq{c}\label{eq:vanilla-model-subst}.\\ % SUBSTITUTION (c1->c2)
P_{edit},                    & \mkern10mu\text{if}\ c\myneq\varepsilon\ \text{and}\ \tilde{c}\myeq\varepsilon.\label{eq:vanilla-model-del}\\ % DELETION (c->e)
1{-}2P_{edit},    & \mkern10mu\text{if}\ c{\myneq}\varepsilon\ \text{and}\ \tilde{c}\myeq{c}\label{eq:vanilla-model-no-change}. % NO-CHANGE (c->c)
\end{empheq}
\end{subequations}

\noindent
It consists of the following components:
\begin{enumerate}[label=(\alph*)]

\item The \emph{insertion probability} $P_{ins}(\tilde{c}\given{\varepsilon})$ in \cref{eq:vanilla-model-ins}.
It describes how likely it is to insert a non-empty character $\tilde{c}\myneq\varepsilon$ and it is uniform over the set of all characters from the alphabet $\Sigma$, except the $\varepsilon$ symbol.

\item The \emph{keep $\varepsilon$ probability} $P_{keep}(\varepsilon\given{\varepsilon})$ in \cref{eq:vanilla-model-ins-no-change}.  %($\varepsilon\rightarrow\varepsilon$)

\item The \emph{substitution probability} $P_{subst}(\tilde{c}\given{c})$ in \cref{eq:vanilla-model-subst}.
It is uniform over the set of all characters from the alphabet $\Sigma$, except the source character $c$ and the $\varepsilon$ symbol.

\item The \emph{deletion probability} $P_{del}(\varepsilon\given{c})$ in \cref{eq:vanilla-model-del}.%($c\rightarrow\varepsilon$)

\item The \emph{keep probability} $P_{keep}(c\given{c})$ in \cref{eq:vanilla-model-no-change}. %($c\rightarrow{c}$)
\end{enumerate}

\noindent
\Cref{eq:vanilla-model-ins,eq:vanilla-model-ins-no-change} correspond to the row in the character confusion matrix $\Gamma$, where $c\myeq\varepsilon$ and form a valid probability distribution: 
\begin{align*}
P_{keep}(\varepsilon\given{\varepsilon}) + \smashoperator[lr]{\sum_{\tilde{c}\,\in\,\Sigma{\setminus}\{\varepsilon\}}}P_{ins}(\tilde{c}\given{c}) &= 1.
\end{align*}
\noindent
Similarly, \cref{eq:vanilla-model-subst,eq:vanilla-model-del,eq:vanilla-model-no-change} correspond to the rows in the character confusion matrix $\Gamma$, where $c\in\Sigma{\setminus}\{\varepsilon\}$, and are also valid probability distributions:
\begin{align*}
P_{del}(\varepsilon\given{c}) + P_{keep}(c\given{c}) + \smashoperator[lr]{\sum_{\tilde{c}\,\in\,\Sigma{\setminus}\{c,\,\varepsilon\}}}P_{subst}(\tilde{c}\given{c}) &= 1
\end{align*}

Finally, for comparison, we present visualizations of the confusion matrices used in our vanilla (\Cref{tab:cmx-vanilla}) and OCR error models (\Cref{tab:cmx-ocr}). 

\begin{figure*}[t]
\centering
\begin{subfigure}[t]{0.9\textwidth}
\centering
\includegraphics[width=0.7\textwidth]{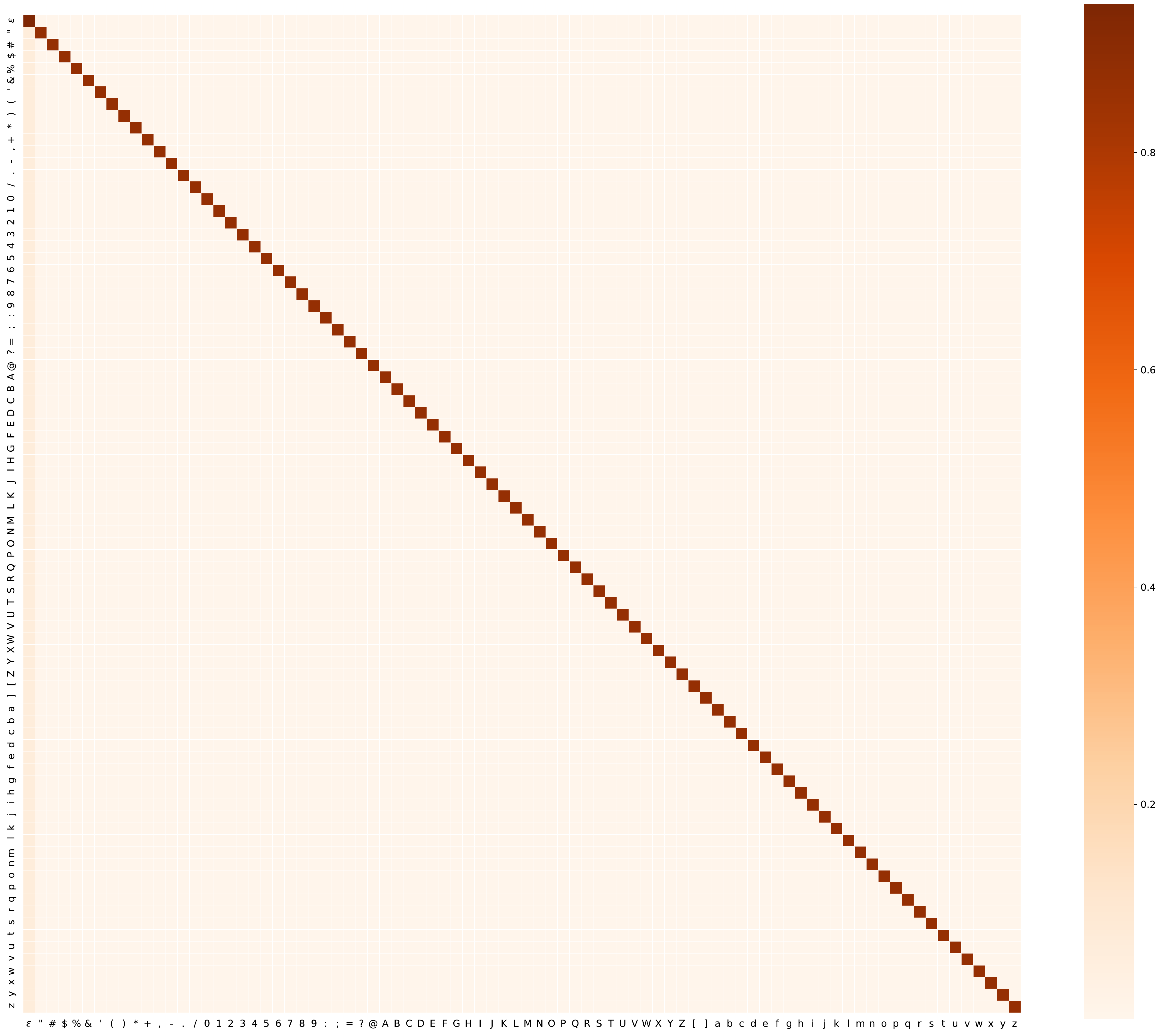}
\caption{Vanilla error distribution used at training time ($\eta=20\%$).}
\label{tab:cmx-vanilla}
\end{subfigure}
\par\smallskip % force a bit of vertical whitespace
\begin{subfigure}[t]{0.9\textwidth}
\centering
\includegraphics[width=0.7\textwidth]{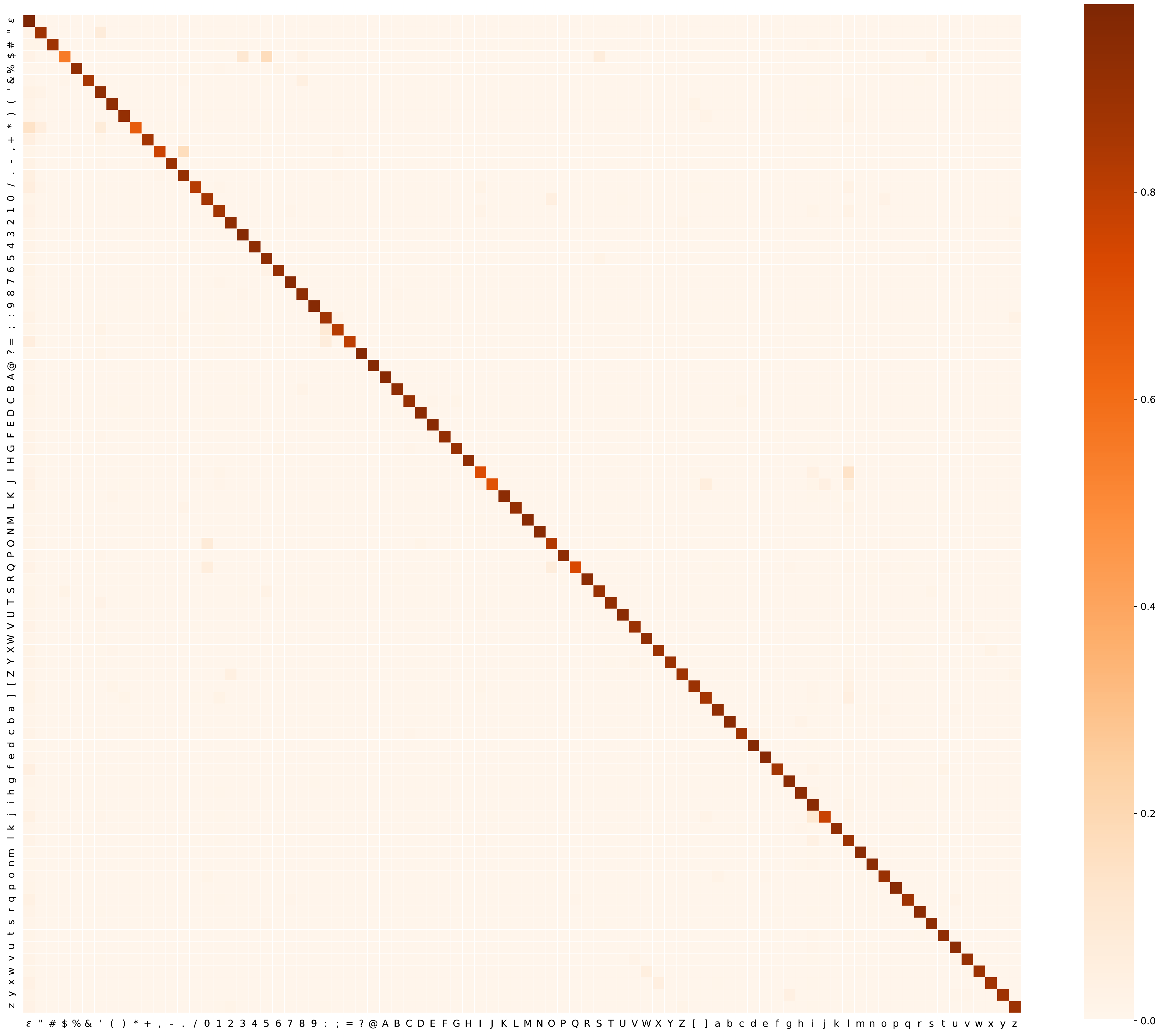}
\caption{Real error distribution estimated from a large document corpus using the Tesseract OCR engine.}
\label{tab:cmx-ocr}
\end{subfigure}
\caption{Confusion matrices for the vanilla and the OCR error distributions. 
Each cell represents $P(\tilde{c}\given{c})$.
The rows correspond to the original characters $c$ and the columns represent the perturbed characters $\tilde{c}$.
In this example, we include all symbols from the alphabet of the English CoNLL 2003 data set. 
The vanilla noise model assigns equal probability to all substitution errors, while the OCR error model is biased towards substitutions of characters with similar shapes like "{\tt I}"$\rightarrow$"{\tt l}", "{\tt \$}"$\rightarrow$"{\tt 5}", "{\tt O}"$\rightarrow$"{\tt 0}" or "{\tt ,}"$\rightarrow$"{\tt .}".
Moreover, the vanilla model assumes that the deletion of a character $c$ is as likely as the sum of substitution probabilities with all non-empty symbols: $P_{del}(\varepsilon\given{c})\myeq\sum_{{\tilde{c}\,\in\,\Sigma{\setminus}\{\varepsilon\}}} P_{subst}(\tilde{c}\given{c})$.
}
\label{tab:cmx}
\end{figure*}

% ====================== Extended Sensitivity Analysis ======================
\subsection{Sensitivity Analysis}
\label{sec:eval2-ext}

In this section, we present the extended version of our sensitivity study\notinsubfile{ (\S\ref{ssec:eval2})}. \Cref{fig:eval2-ext} summarizes the results on the synthetic data distribution with various test- and training-time noise levels ($\eta_{test}$ and $\eta_{train}$, respectively) and weighting factors $\alpha$. We noticed a similar trend as in our initial analysis. As the level of noise $\eta_{test}$ increases, the overall accuracy decreases, but this trend is less pronounced for $\alpha \neq 0$. At the same time, the gap between the models trained with and without our auxiliary objectives becomes larger. 

\begin{figure*}[htbp]
\centering
\begin{subfigure}[t]{0.49\textwidth}
\centering
\includegraphics[height=3.6cm]{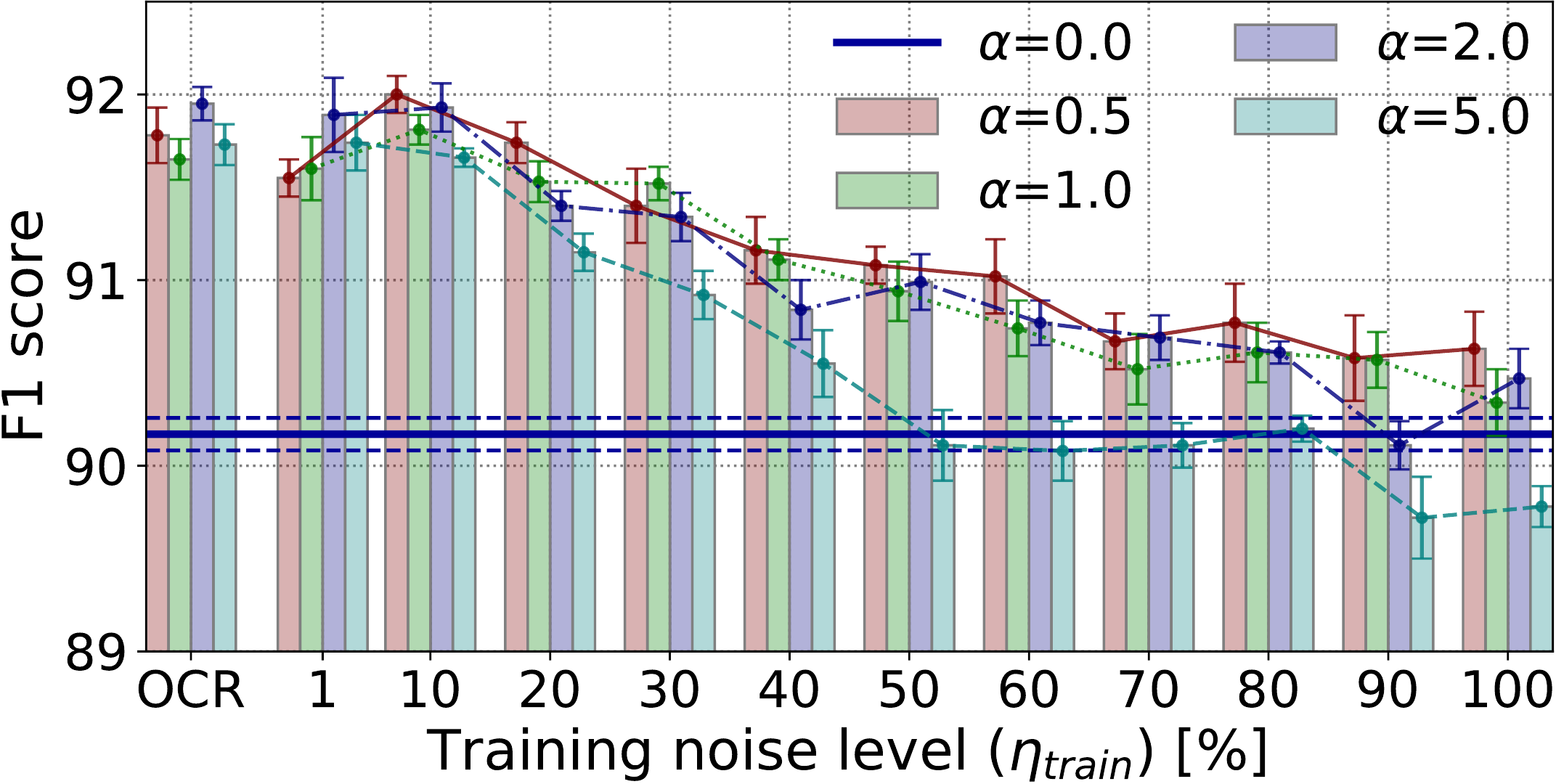}
\caption{Data augmentation objective (synthetic noise: $\eta_{test}\myeq1\%$)}
\end{subfigure}%
%~
\begin{subfigure}[t]{0.49\textwidth}
\centering
\includegraphics[height=3.6cm]{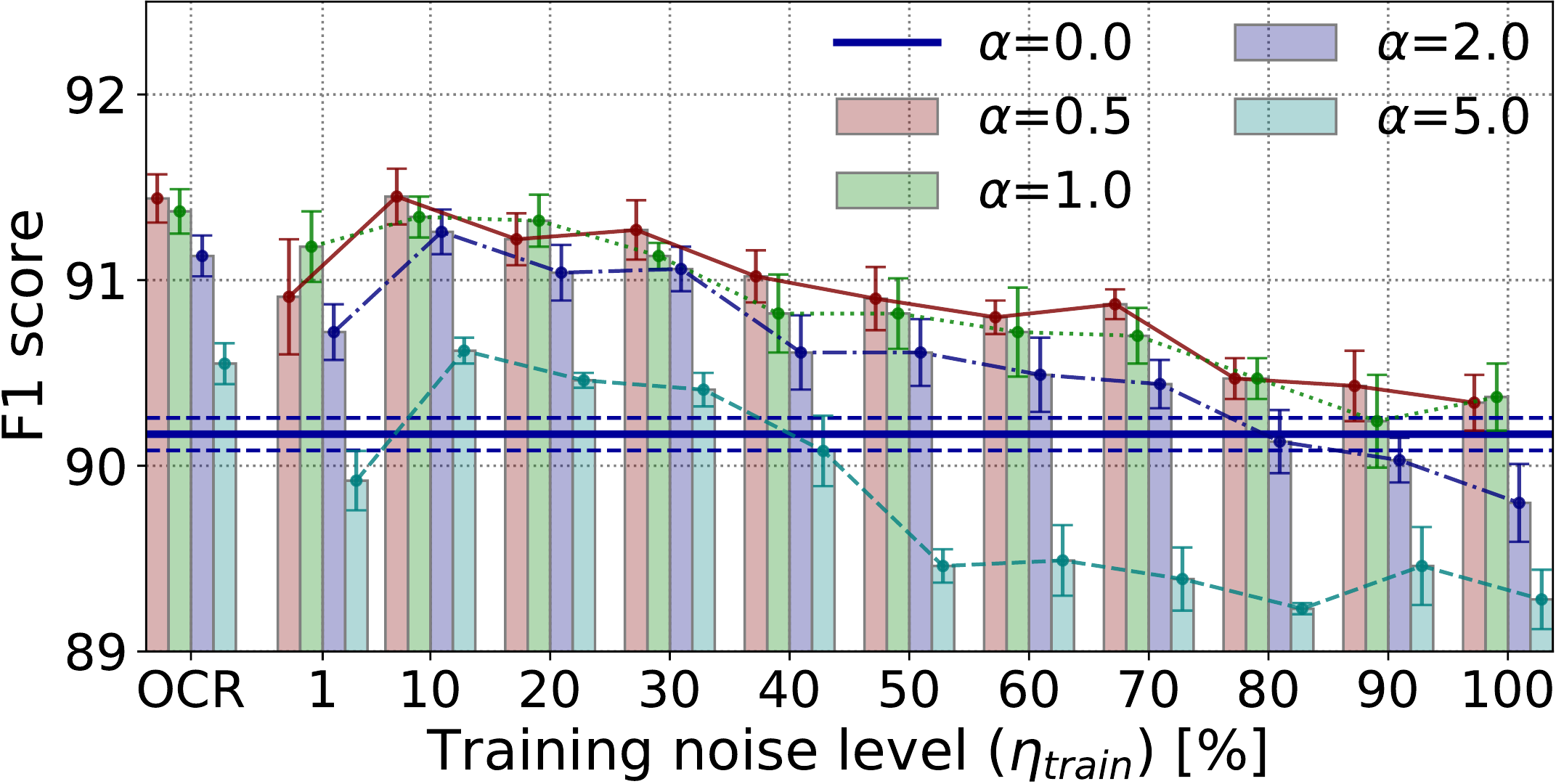}
\caption{Stability training (synthetic noise: $\eta_{test}\myeq1\%$)}
\end{subfigure}
\par\smallskip % force a bit of vertical whitespace
\begin{subfigure}[t]{0.49\textwidth}
\centering
\includegraphics[height=3.6cm]{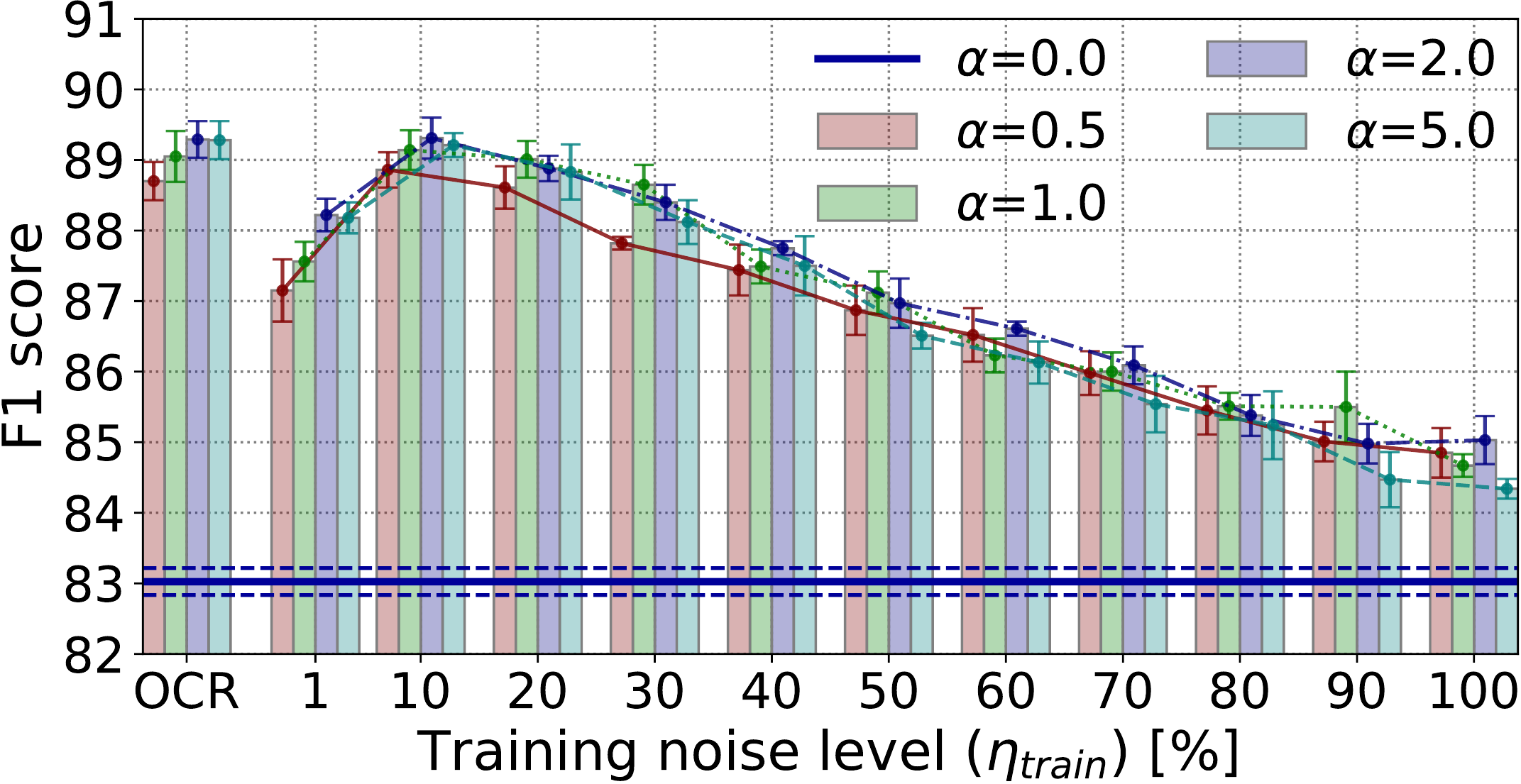}
\caption{Data augmentation objective (synthetic noise: $\eta_{test}\myeq5\%$)}
\end{subfigure}
%~ 
\begin{subfigure}[t]{0.49\textwidth}
\centering
\includegraphics[height=3.6cm]{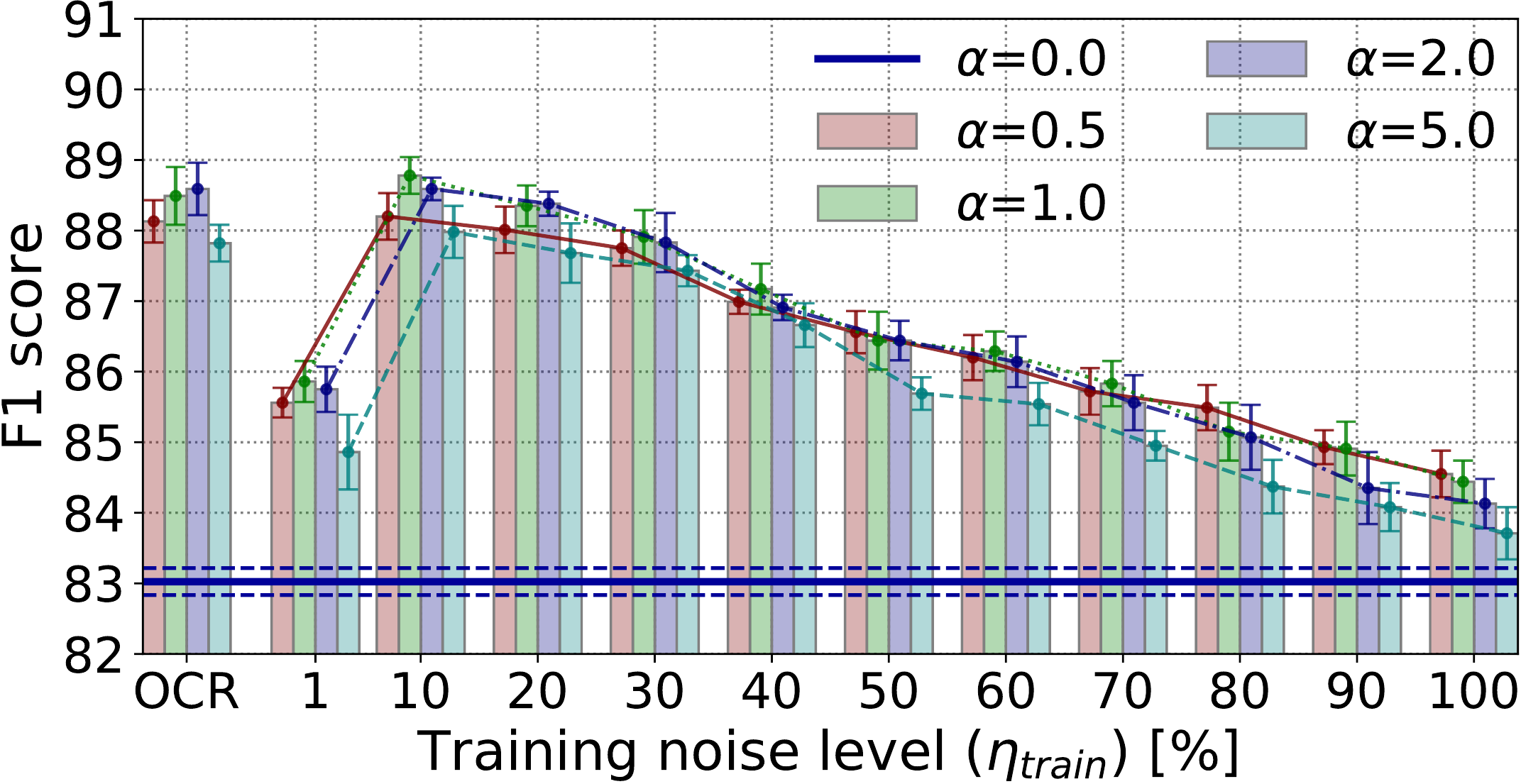}
\caption{Stability training  objective(synthetic noise: $\eta_{test}\myeq5\%$)}
\end{subfigure}  
\par\smallskip % force a bit of vertical whitespace  
\begin{subfigure}[t]{0.49\textwidth}
\centering
\includegraphics[height=3.6cm]{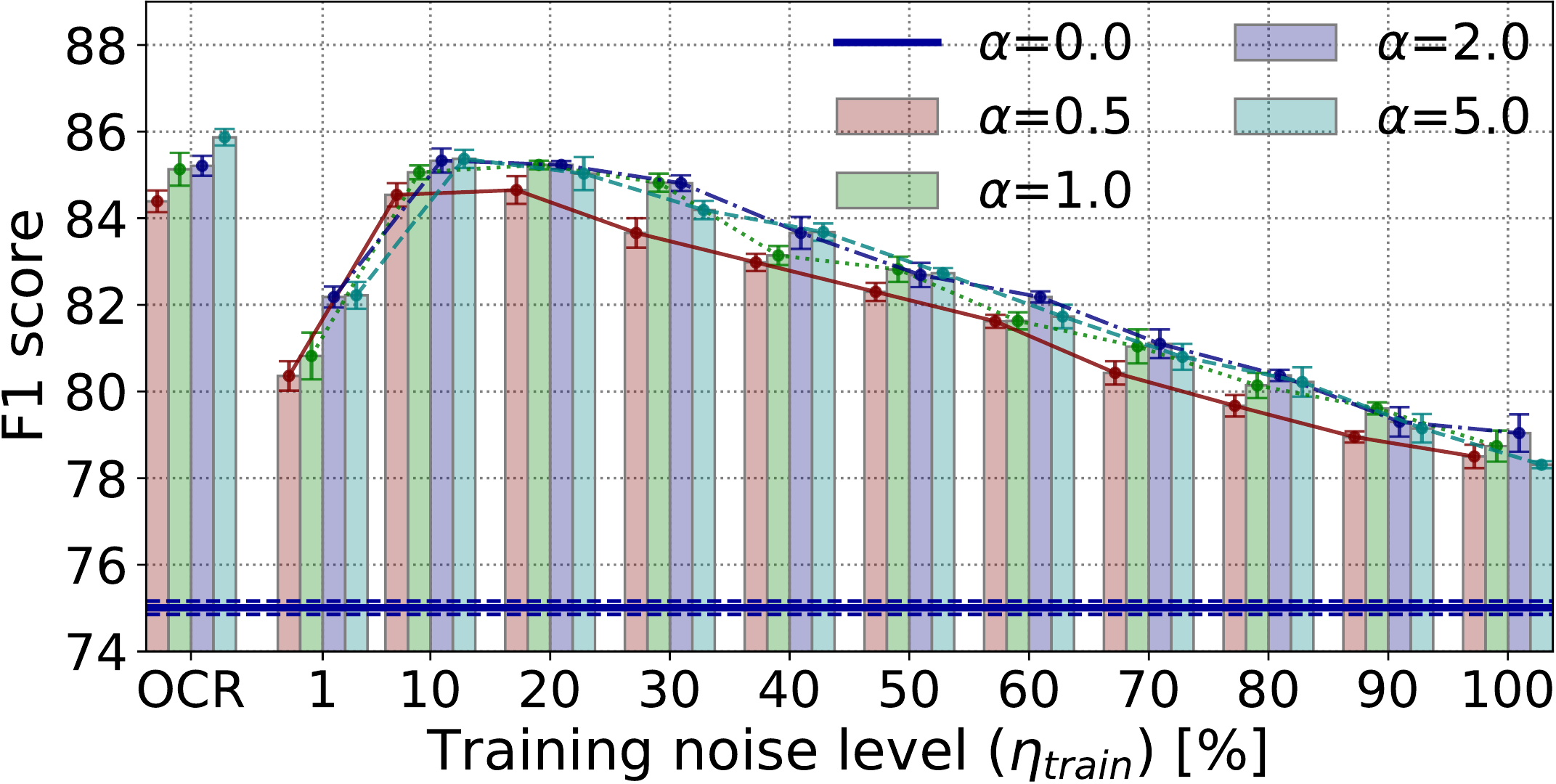}
\caption{Data augmentation objective (synthetic noise: $\eta_{test}\myeq10\%$)}
\end{subfigure}
%~ 
\begin{subfigure}[t]{0.49\textwidth}
\centering
\includegraphics[height=3.6cm]{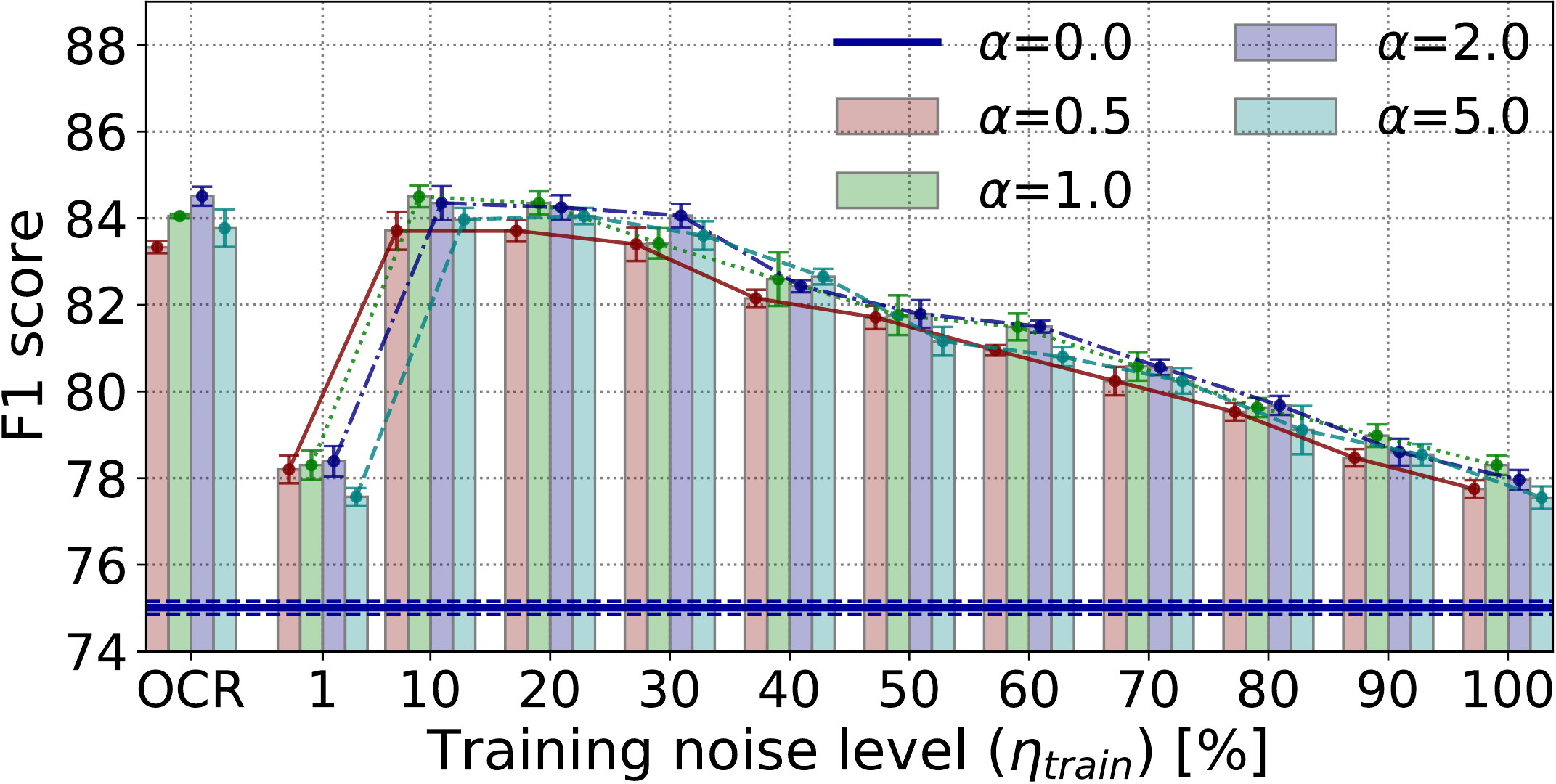}
\caption{Stability training objective (synthetic noise: $\eta_{test}\myeq10\%$)}
\end{subfigure}  
\par\smallskip % force a bit of vertical whitespace
\begin{subfigure}[t]{0.49\textwidth}
\centering
\includegraphics[height=3.6cm]{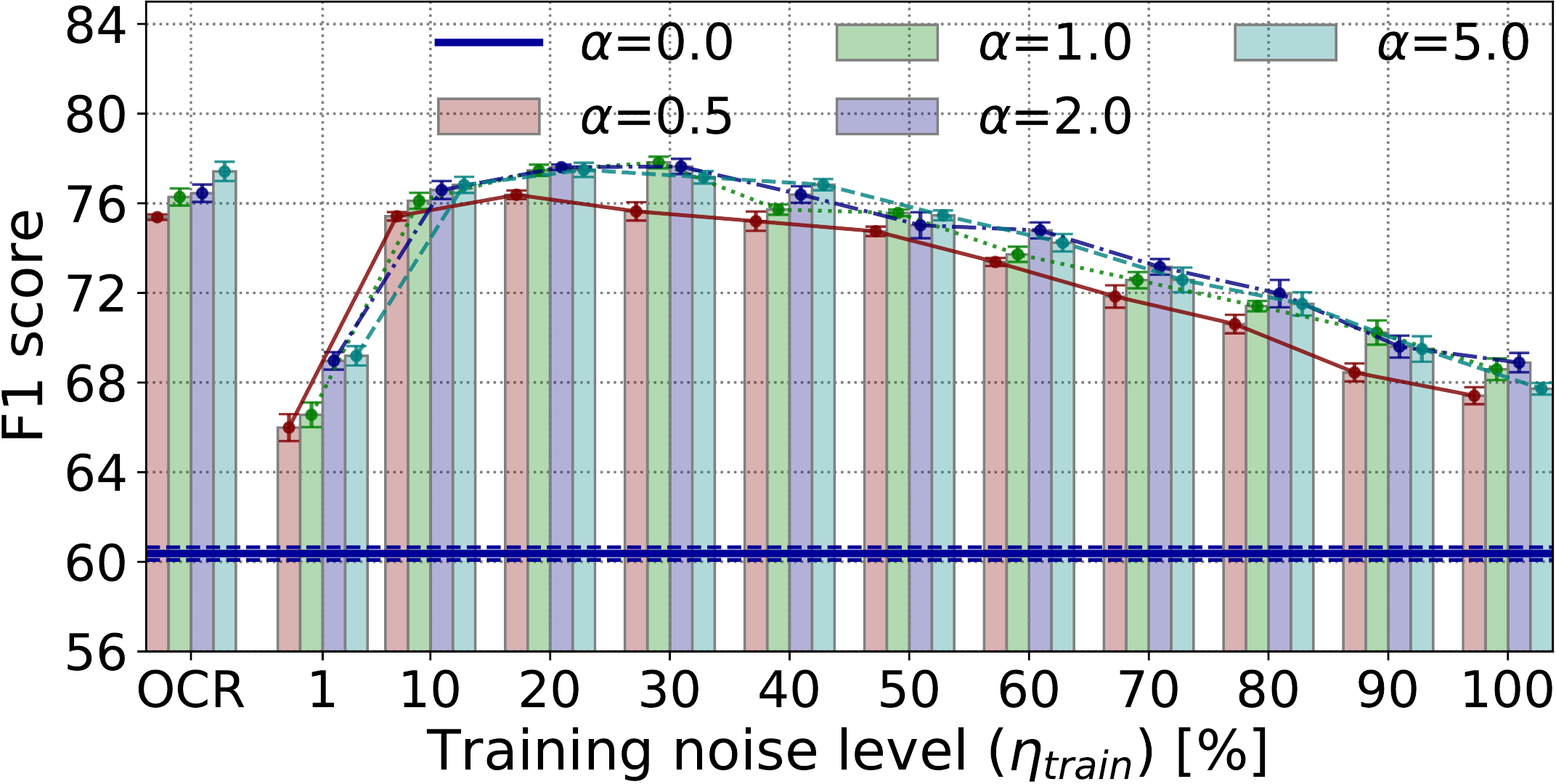}
\caption{Data augmentation objective (synthetic noise: $\eta_{test}\myeq20\%$)}
\end{subfigure}
%~ 
\begin{subfigure}[t]{0.49\textwidth}
\centering
\includegraphics[height=3.6cm]{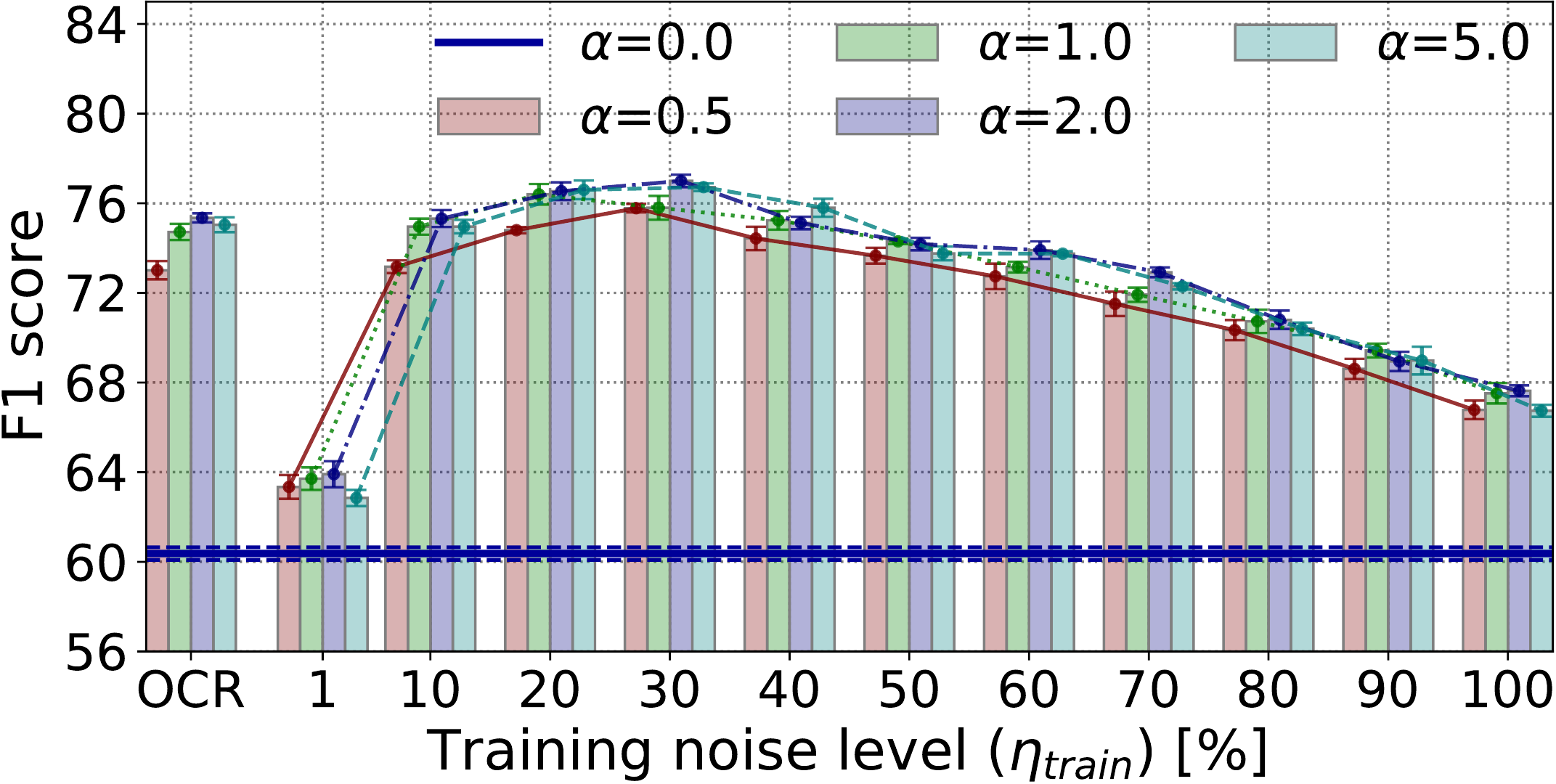}
\caption{Stability training objective (synthetic noise: $\eta_{test}\myeq20\%$)}
\end{subfigure}  
~  
\caption{Extended results of our sensitivity analysis on the English CoNLL 2003 test data (\S\ref{sec:eval2-ext}). Each figure presents the results of models trained using one of our auxiliary training objectives on the original data perturbed with various levels of synthetic noise. The bar marked as "OCR" represents a model trained using the OCR noise distribution. Other bars correspond to models trained using synthetic noise distribution and different hyper-parameters ($\alpha$, $\eta_{train}$).}
\label{fig:eval2-ext}
\end{figure*}

% ====================== Qualitative Analysis ======================

\subsection{Qualitative Analysis}
\label{ssec:eval-qualitative}

In this section, we compared the outputs generated by the baseline models trained with and without our auxiliary training objectives (\Cref{tab:output}). We found that the NAT method improved robustness to capitalization errors (the first and the fourth row in \Cref{tab:output-typos}) as well as to substitutions (the second, the third and the fifth row in \Cref{tab:output-typos} and the first, the second, the fourth and the fifth row in \Cref{tab:output-ocr}), deletions (the fifth row in \Cref{tab:output-typos}) and insertions of characters (the third and the fifth row in \Cref{tab:output-ocr}). Moreover, it better recognized the semantics of the sentence in the third row of \Cref{tab:output-typos}, where the location name was creatively rewritten (\textit{Brazland} instead of \textit{Brazil}).

{\setlength{\tabcolsep}{0pt}\renewcommand{\arraystretch}{1.0}
\begin{table*}[htbp]
\begin{subtable}{\textwidth}\centering\small
\begin{tabular}{L{0.04}L{0.19}L{0.77}}
\toprule
\multirow{3}{*}{1.}
& Reference result& {\small\tt 7-1 Raul <B-PER> Gonzalez <E-PER> 7-1 Juan <B-PER> Pizzi <E-PER>}\\
& NAT output      & {\small\tt 7-1 raul <B-PER> gonzalez <E-PER> 7-1 juan <B-PER> Pizzi <E-PER>}\\
& Baseline output & {\small\tt 7-1 raul gonzalez <S-PER> 7-1 juan Pizzi <S-PER>}\\
\midrule
\multirow{3}{*}{2.}
& Reference result& {\small\tt 6. Heidi <B-PER> Zurbriggen <E-PER> ( Switzerland <S-LOC> ) 153}\\
& NAT output      & {\small\tt 6. Heidi <B-PER> Zurbriggen <E-PER> ( swizzerland <S-LOC> ) 153}\\
& Baseline output & {\small\tt 6. Heidi <B-PER> Zurbriggen <E-PER> ( swizzerland ) 153}\\
%\midrule
%Reference result& {\small\tt Damascus <S-LOC> denies aiding the rebels .}\\
%NAT output      & {\small\tt Damascuse <S-LOC> denies aiding de rebels .}\\
%Baseline output & {\small\tt Damascuse <S-PER> denies aiding de rebels .}\\
\midrule
\multirow{3}{*}{3.}
& Reference result& {\small\tt Plastic surgery gets boost in Brazil <S-LOC> .}\\
& NAT output      & {\small\tt Plastic surgury hets boost is Brazland <S-LOC> .}\\
& Baseline output & {\small\tt Plastic surgury hets boost is Brazland <S-PER> .}\\
\midrule
\multirow{3}{*}{4.}
& Reference result& {\small\tt Waltraud <B-PER> Zimmer <E-PER> , Rödermark-Ober-Roden <S-LOC>}\\
& NAT output      & {\small\tt Waltraud <B-PER> zimmer <E-PER> , Rödermark-Ober-Roden <S-LOC>}\\
& Baseline output & {\small\tt Waltraud <S-PER> zimmer , Rödermark-Ober-Roden <S-LOC>}\\
\midrule
\multirow{3}{*}{5.}
& Reference result& {\small\tt Deutschland <S-LOC> ist noch nicht Teil der Reiseroute . "}\\
& NAT output      & {\small\tt Deutshland <S-LOC> is nach nich Teil der Reiseroute . "}\\
& Baseline output & {\small\tt Deutshland <S-PER> is nach nich Teil der Reiseroute . "}\\
%\midrule
%Reference result& {\small\tt Frank-Walter <B-PER> Steinmeier <E-PER> muss die schlechte Nachricht überbringen .}\\
%NAT output      & {\small\tt Frank-Walter <B-PER> Steinmeier <E-PER> muß die schlechte Naricht überbringen .}\\
%Baseline output & {\small\tt Frank-Walter <B-PER> Steinmeier <E-PER> muß die schlechte Naricht <S-PER> überbringen .}\\
%\midrule
%Reference result& {\small\tt Auch für sie kostet die Bundesliga <S-ORG> 14,90 Euro <S-OTH> im Monat .}\\
%NAT output      & {\small\tt Aauch fur si kosstet di Bundesliga <S-ORG> 14,90 Euro <S-OTH> im Monat .}\\
%Baseline output & {\small\tt Aauch <B-OTH> fur <I-OTH> si <E-OTH> kosstet di Bundesliga 14,90 Euro <S-OTH> im Monat .}\\
\bottomrule
\end{tabular}
\caption{Misspellings.}
\label{tab:output-typos}
\end{subtable}
\par\smallskip % force a bit of vertical whitespace
\begin{subtable}{\textwidth}\centering\small
\begin{tabular}{L{0.03}L{0.2}L{0.77}}
\toprule
\multirow{3}{*}{1.}
& Reference result& {\small\tt Hapoel <B-ORG> Jerusalem <E-ORG> 12 4 1 7 10 18 13}\\
& NAT output      & {\small\tt Hapoel <B-ORG> lerusalem <E-ORG> I2 A 1 7 10 18 13}\\
& Baseline output & {\small\tt Hapoel <S-ORG> lerusalem I2 A 1 7 10 18 13}\\
\midrule
\multirow{3}{*}{2.}
& Reference result& {\small\tt SOCCER - SPANISH <S-MISC> FIRST DIVISION RESULT / STANDINGS .}\\
& NAT output      & {\small\tt SOCCER - SPANlSH <S-MISC> FIRST DIVISiOW RESULT / STA'DINGS .}\\
& Baseline output & {\small\tt SOCCER - SPANlSH <S-PER> FIRST DIVISiOW RESULT / STA'DINGS .}\\
\midrule
\multirow{3}{*}{3.}
& Reference result& {\small\tt EU <S-ORG> , Poland <S-LOC> agree on oil import tariffs .}\\
& NAT output      & {\small\tt EU <S-ORG> , Po'land <S-LOC> agree on oil import tarifs .}\\
& Baseline output & {\small\tt EU <S-ORG> , Po'land <S-ORG> agree on oil import tarifs .}\\
\midrule
\multirow{3}{*}{4.}
& Reference result& {\small\tt Schlamm scheint zu helfen - Yahoo <B-ORG> ! <E-ORG>}\\
& NAT output      & {\small\tt Schlamm scheint zu helfen - Yaho0 <B-ORG> ! <E-ORG>}\\
& Baseline output & {\small\tt Schlamm scheint zu helfen - Yaho0 <S-PER> !}\\
\midrule
\multirow{3}{*}{5.}
& Reference result& {\small\tt Fachverband <B-ORG> für <I-ORG> Hauswirtschaft <E-ORG> :}\\
& NAT output      & {\small\tt Fachverbandi <B-ORG> für <I-ORG> Hauswi'tschaTt <E-ORG> :}\\
& Baseline output & {\small\tt Fachverbandi für Hauswi'tschaTt :}\\
%\midrule
%Reference text  & {\small\tt }\\
%NAT output      & {\small\tt }\\
%Baseline output & {\small\tt }\\
\bottomrule
\end{tabular}
\caption{OCR errors.}
\label{tab:output-ocr}
\end{subtable}
\caption{Outputs produced by the models trained with and without our auxiliary NAT objectives ({\it NAT output} and {\it Baseline output}, respectively). We demonstrate examples that contain misspellings and OCR errors, where the models trained with the auxiliary NAT objectives correctly recognized all tags, while the baseline models either misclassified or completely missed some entities.
}
\label{tab:output}
\end{table*}}

% ====================== Hyper-parameters ======================
\section{Hyper-parameters}
\label{sec:hyperparams}

We present the detailed hyper-parameters of the sequence labeling model $f(x)$ used in our experiments\notinsubfile{ (\S\ref{sec:experiments})}. Note that dropout was applied both before and after the LSTM layer (\Cref{tab:hyperparams-sequence-labeling}).

{\setlength{\tabcolsep}{5pt}\renewcommand{\arraystretch}{1.0}
%https://texblog.org/2019/06/03/control-the-width-of-table-columns-tabular-in-latex/
\begin{table}[H]
\centering\small
\begin{tabular}{X{0.55}Y{0.35}}
\toprule
Parameter name & Parameter value \\
\midrule
Tagging schema & BIOES \\
Mini batch size & 32 \\
Max. epochs & 100 \\
LSTM \# hidden layers & 1 \\
LSTM \# hidden units & 256 \\
Optimizer & SGD \\
Initial learning rate & 0.1 \\
Learning rate anneal factor & 0.5\\
Minimum learning rate & 0.0001 \\
Word dropout level & 0.05 \\
Variational dropout level & 0.5\\
Patience & 3\\
%\dots & \dots\\
\bottomrule
\end{tabular}
\caption{Hyper-parameters of the sequence labeling model $f(x)$ used in our experiments.}
\label{tab:hyperparams-sequence-labeling}
\end{table}}

% ====================== Data Set Statistics and Estimated Error Rates ======================

\section{Data Set Statistics and Estimated Error Rates }
\label{sec:datasets}

In this section, we present the detailed statistics of the data sets used in our NER experiments (\Cref{tab:datasets-ner}). Following~\citet{akbik-etal-2018-contextual}, we used the revisited version of German CoNLL 2003, which was prepared in 2006 and is believed to be more accurate, as the previous version was done by non-native speakers\footnote{The revisited annotations are available on the official website of the CoNLL 2003 shared task: \url{https://www.clips.uantwerpen.be/conll2003/ner/}.}. Moreover, we used only the inner layer of annotation for GermEval 2014.

Finally, in \Cref{tab:error-rates}, we report estimated error rates for all data sets and all noising procedures used in our experiments.

{\setlength{\tabcolsep}{4pt}\renewcommand{\arraystretch}{1.0}
\begin{table}[htbp]
\begin{subtable}{\columnwidth}\centering\small
\begin{tabular}{lrrrr}
\toprule
          & Train & Dev & Test & Total\\
\midrule
Sentences & 14,041  & 3,250  & 3,453  & 20744 \\
Tokens    & 203,621 & 51,362 & 46,435 & 301418 \\
%\midrule
PER       & 6,600   & 1,842  & 1,617  & 10059 \\
LOC       & 7,140   & 1,837  & 1,668  & 10645 \\
ORG       & 6,321   & 1,341  & 1,661  & 9323 \\
MISC      & 3,438   & 922   & 702     & 5062 \\
\bottomrule
\end{tabular}
\caption{English CoNLL 2003.}
\end{subtable}
\par\smallskip % force a bit of vertical whitespace
\begin{subtable}{\columnwidth}\centering\small
\begin{tabular}{lrrrr}
\toprule
          & Train & Dev & Test & Total \\
\midrule
Sentences & 12,705  & 3,068  & 3,160  & 18933 \\
Tokens    & 207,484 & 51,645 & 52,098 & 311227 \\
%\midrule
PER       & 2,801   & 1,409  & 1,210  & 5420 \\
LOC       & 4,273   & 1,216  & 1,051  & 6540 \\
ORG       & 2,154   & 1,090  & 584    & 3828 \\
MISC      & 780     & 216    & 206    & 1202 \\
\bottomrule
\end{tabular}
\caption{German CoNLL 2003.}
\end{subtable}
\par\smallskip % force a bit of vertical whitespace
\begin{subtable}{\columnwidth}\centering\small
\begin{tabular}{lrrrr}
\toprule
          & Train & Dev & Test & Total \\
\midrule
Sentences    & 24,000  & 2,200  & 5,100  & 31300 \\
Tokens       & 452,853 & 41,653 & 96,499 & 591005 \\
%\midrule
PER          & 7,679   & 711 & 1,639     & 10029 \\
PER-deriv    & 62      & 2 & 11          & 75 \\ 
PER-part     & 184     & 18 & 44         & 246 \\
LOC          & 8,281   & 763 & 1,706     & 10750 \\
LOC-deriv    & 2,808   & 235 & 561       & 3604 \\
LOC-part     & 513     & 52 & 109        & 674 \\
ORG          & 5,255   & 496 & 1,150     & 6901 \\
ORG-deriv    & 41      & 3 & 8           & 52 \\
ORG-part     & 805     & 91 & 172        & 1068 \\
MISC         & 3,024   & 269 & 697       & 3990 \\
MISC-deriv   & 236     & 16 & 39         & 291 \\
MISC-part    & 190     & 18 & 42         & 250 \\
\bottomrule
\end{tabular}
\caption{GermEval 2014.}
\end{subtable}
\caption{Statistics of the data sets used in our NER experiments\notinsubfile{ (\S\ref{sec:experiments})}. We present statistics of the training (Train) development (Dev) and test (Test) sets, including the number of sentences, tokens, and entities: person names (PER), locations (LOC), organizations (ORG) and miscellaneous (MISC). 
The GermEval 2014 data set defines two additional fine-grained sub-labels: \mbox{"-part"} and \mbox{"-deriv"} that mark derivation and compound words, respectively, which stand in direct relation to Named Entities.
}
% the original CoNLL2003 paper report a wrong number of sentences, see: https://github.com/davidsbatista/NER-datasets/issues/2
%"What I found out that the published CoNLL03 dataset considers -DOCSTART as a sentence instance. So mostly papers in the NER report number from the original paper!"
\label{tab:datasets-ner}
\end{table}}

{\renewcommand{\arraystretch}{1.0}\setlength{\tabcolsep}{2pt}
\begin{table}[htbp]
\begin{subtable}{\columnwidth}\centering\small
\begin{tabular}{X{0.35}*{3}{Z{0.18}}}
\toprule
& {OCR noise}  & {Mis-spellings$^\dag$} & {Mis-spellings$^\ddag$} \\
\midrule
{English\,CoNLL\,2003} & 8.9\% & 16.5\% & 9.8\%  \\
{German\,CoNLL\,2003}  & 9.0\% & 8.3\%  & 8.0\%  \\
{GermEval 2014}        & 9.3\% & 8.6\%  & 8.2\%  \\
\bottomrule
\end{tabular}
\caption{Character Error Rates.}
%\label{tab:error-rates-cer}
\end{subtable}
\quad
%\par\smallskip % force a bit of vertical whitespace
\begin{subtable}{\columnwidth}\centering\small
\begin{tabular}{X{0.35}*{3}{Z{0.18}}}
\toprule
& {OCR noise}  & {Mis-spellings$^\dag$} & {Mis-spellings$^\ddag$} \\
\midrule
{English\,CoNLL\,2003} & 35.6\% & 55.4\% & 48.3\% \\
{German\,CoNLL\,2003}  & 39.5\% & 26.5\% & 45.5\% \\
{GermEval 2014}        & 41.2\% & 27.0\% & 47.9\% \\
\bottomrule
\end{tabular}
\caption{Word Error Rates.}
%\label{tab:error-rates-wer}
\end{subtable}
\caption{
%Estimated error rates of text produced using different noise distributions.
Error rate estimation for different noise distributions.
OCR noise is modeled with the character confusion matrix, whereas misspellings are induced using look-up tables released by \citet{DBLP:conf/iclr/BelinkovB18}$^\dag$ and \citet{piktus-etal-2019-misspelling}$^\ddag$.}
\label{tab:error-rates}
\end{table}}

\onlyinsubfile{
\bibliographystyle{acl_natbib}
\bibliography{anthology,paper-acl2020}
}

\fi

\end{document}